\documentclass[letterpaper]{article} 
\usepackage{aaai2026}  
\usepackage{times}  
\usepackage{helvet}  
\usepackage{courier}  
\usepackage[hyphens]{url}  
\usepackage{graphicx} 
\usepackage{natbib}  
\usepackage{caption} 
\usepackage[utf8]{inputenc} 

\usepackage{url}            
\usepackage{booktabs}       
\usepackage{amsfonts}       
\usepackage{nicefrac}       
\usepackage{microtype}      
\usepackage{xcolor}         
\usepackage{multirow}
\usepackage{adjustbox}
\usepackage{amsmath,amssymb}
\usepackage{multirow}
\usepackage{tikz}
\usepackage{subcaption}
\usepackage[font=small]{caption}
\usepackage[super]{nth}
\usepackage{algorithm}
\usepackage{algpseudocode}
\usepackage{tabularx}
\usepackage{makecell}
\usepackage{enumitem}
\usepackage{bbding}
\usepackage{threeparttable}
\usepackage{colortbl}
\usepackage{graphicx}

\graphicspath{ {./figures/}, {./figures/image_experiment/}, {./figures/tex/}, {./figures/supp/}, {./figures/analysis/} }
\def\paperTitle{CorrectAD: A Self-\underline{Correct}ing Agentic System to \\ Improve End-to-end Planning in \underline{A}utonomous \underline{D}riving}

\newcommand{\mypara}[1]{\vspace{1mm}\noindent\textbf{#1}~~}

\title{\paperTitle}

\author{
    Enhui Ma\textsuperscript{\rm 1}\thanks{Co-first authors}\thanks{Work done during an internship at Li Auto Inc.},
    Lijun Zhou\textsuperscript{\rm 2}\footnotemark[1],
    Tao Tang\textsuperscript{\rm 2}\footnotemark[2],
    Jiahuan Zhang\textsuperscript{\rm 1}\footnotemark[3],
    Junpeng Jiang\textsuperscript{\rm 2}\footnotemark[2],
    Zhan Zhang\textsuperscript{\rm 1}\thanks{Work done during their visiting at Autolab, Westlake University.},\\
    Dong Han\textsuperscript{\rm 1}\footnotemark[3],
    Kun Zhan\textsuperscript{\rm 2},
    Xueyang Zhang\textsuperscript{\rm 2},
    Xianpeng Lang\textsuperscript{\rm 2},
    Haiyang Sun\textsuperscript{\rm 2},
    Xia Zhou\textsuperscript{\rm 2},
    Di Lin\textsuperscript{\rm 3},
    Kaicheng Yu\textsuperscript{\rm 1}\thanks{Corresponding Author}
}

\affiliations{
    \textsuperscript{\rm 1}Autolab, Westlake University \textsuperscript{\rm 2}Li Auto Inc. 
    \textsuperscript{\rm 3}Tianjin University\\
    \{maenhui, kyu\}@westlake.edu.cn
}

\begin{document}

\maketitle

\begin{figure*}[ht!]
\centering
\vspace{-0.4cm}
\includegraphics[width=\linewidth]{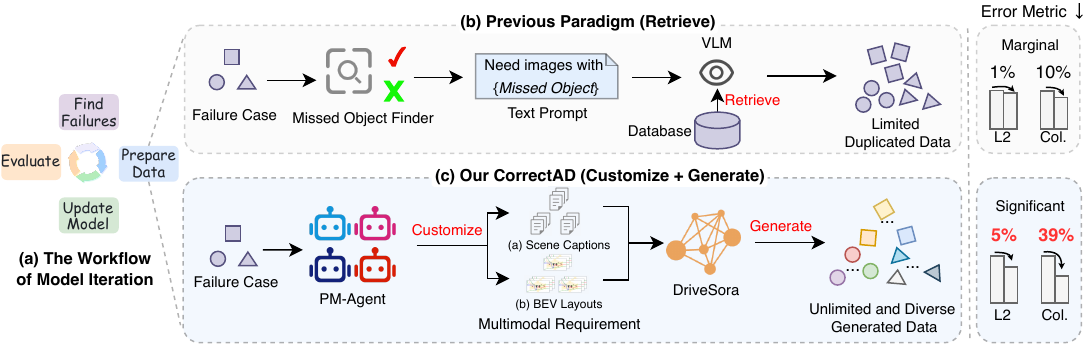}
\vspace{-0.4cm}
\caption{\textbf{(a)}: The workflow of one model iteration consists of 4 steps: finding failure cases, preparing training data, model updating, followed by evaluation and iteration again. \textbf{The key issue is how to prepare specific training data to correct the failure cases.}
\textbf{(b)}: 
Previous paradigm was retrieval-based, i.e., retrieving similar data from the existing dataset and auto-labeling them, which severely limits the diversity of training data.
\textbf{(c)}: Our proposed agentic system, \textbf{CorrectAD}, is custom-generated. 
We first propose \textbf{PM-Agent}, similar to the role of Product Manager, to formulate data requirements by analyzing failure cases. Then, we propose a generative model \textbf{DriveSora}, similar to the role of Data Department, to generate high-fidelity training data aligned with the data requirements requested by PM-Agent.
Our approach outperforms previous methods in L2 and collision rate (Col.) for end-to-end planning models.
}
\label{fig:teaser}
\vspace{-0.2cm}
\end{figure*}

\begin{abstract}
End-to-end planning methods are the de-facto standard of the current autonomous driving system, while the robustness of the data-driven approaches suffers due to the notorious ``long-tail" problem (i.e., rare but safety-critical failure cases). 
In this work, we explore whether recent diffusion-based video generation methods (a.k.a. world models), paired with structured 3D layouts, can enable a fully automated pipeline to self-correct such failure cases.  
We first introduce an agent to simulate the role of product manager, dubbed \textbf{PM-Agent}, which formulates data requirements to collect data similar to the failure cases. 
Then, we use a generative model that can simulate both data collection and annotation. However, existing generative models struggle to generate high-fidelity data conditioned on 3D layouts. To address this, we propose \textbf{DriveSora}, which can generate spatiotemporally consistent videos aligned with the 3D annotations requested by PM-Agent.
We integrate these components into our self-correcting agentic system, \textbf{CorrectAD}.
Importantly, our pipeline is end-to-end model agnostic and can be applied to improve any end-to-end planner.
Evaluated on both nuScenes and a more challenging in-house dataset across multiple end-to-end planners, CorrectAD corrects 62.5\% and 49.8\% of failure cases, reducing collision rates by 39\% and 27\%, respectively. 

\end{abstract}
\section{Introduction}

End-to-end (E2E) autonomous driving has garnered increasing attention~\cite{hu2023uniad, Jiang2023VADVS, Yang2023VisualPC}, which directly learns to plan motions from raw sensor inputs, thereby reducing heavy reliance on hand-crafted rules and 
avoiding cascading modules.
Deploying robust E2E model is critical for real-world autonomy. However, long-tail scenarios encountered on the road can 
cause catastrophic failures due to limited representation in training data. To adapt to diverse and evolving driving environments, E2E models must be continuously refined. Yet, manually collecting high-quality data for such failure scenarios remains costly and risky, especially for dangerous situations. This problem leads to the emergence of an agentic system that helps E2E models self-correct, keeping them adaptable and effective.

To address this, we draw inspiration from the current data development paradigm of autonomous driving companies, which usually consists of the following steps: product managers receive failure case feedback from the deployment team, then they formulate data requirements and task the data team with collecting and annotating similar scenarios to augment the training set (see Fig.~\ref{fig:teaser}(a)). While effective, this manual process incurs drastically high costs in both data collection and annotation, often taking weeks and thousands of dollars per scenario. Alternative solutions~\cite{su2024aide} (see Fig.~\ref{fig:teaser}(b)) attempt to retrieve and auto-labeling similar data from the existing training dataset, but this severely limits scene diversity and cannot handle unseen failure cases.

In this paper, we propose a fully agentic system to simulate such process towards a self-correcting loop. As illustrated in Fig.~\ref{fig:teaser}(c), to substitute the data department's collection and annotation work, we use a generative model, dubbed as \textbf{DriveSora}, which can simulate the data collection and annotation process by generating multi-view videos controlled by precise 3D scene annotation. 
Unlike prior works that randomly generate scenes~\cite{gao2023magicdrive, wen2023panacea, yang2023bevcontrol}, 
our system focuses on generating targeted data tailored to failure correction.
Yet, the generative model cannot directly take a failure case video to generate such data. To this end, we build an agent to simulate product manager, dubbed \textbf{PM-Agent}. This agent focuses on analyzing failure causes using VLM's reasoning abilities, and then formulates multimodal requirements (including bird's-eye-view layouts and scene descriptions) to interact with the generative model. Finally, 
by incorporating the generated data into the training dataset, our self-correcting agentic system, \textbf{CorrectAD}, significantly improves the robustness of downstream E2E models. Importantly, our approach is agnostic to E2E models and can be applied across diverse planners. 
We demonstrate the effectiveness of CorrectAD on both nuScenes and a challenging in-house dataset, correcting 62.5\% and 49.8\% of failure cases respectively, and reducing collision rates by 39\% and 27\%.
Our contributions can be summarized as follows:

\begin{itemize}[nolistsep, leftmargin=0.5cm]
\setlength{\itemsep}{0.05cm}
\setlength{\parsep}{0pt}
\setlength{\parskip}{0pt}

\item We introduce an agentic system to improve the E2E model by self-correcting failure cases. 
\item We propose PM-Agent that links failure cases and generative model, by analyzing failure causes and formulating multimodal requirements for data generation.
\item We propose DriveSora, a controllable video generation model that surpasses prior works by 10.6\% in FVD and 5.8\% in NDS.
\item We validate CorrectAD across datasets and planners, showcasing its E2E model-agnostic nature and substantial performance gains.

\end{itemize}

\section{Related work}
\label{related}

\mypara{Self-correction in Autonomous Driving.}
Self-correction involves a system detecting its errors and refining its decision-making ability to meet task requirements more effectively~\cite{mitchell2018never, valmeekam2023can}. Vision language models (VLMs), with strong semantic and reasoning abilities, can assist in error validation and correction~\cite{pan2023automatically, madaan2024self, piche2024self,zhang2025srllm,zhang2025ascending,zhang2025webpilot}. In autonomous driving, VLMs have improved decision reliability by providing external feedback to adjust autonomous driving outputs~\cite{fu2024drive, yang2023llm4drive, cui2023drivellm, wen2023dilu}. However, this paradigm does not update the training data within the autonomous driving model, thus not to implement targeted optimizations based on failure cases. 
Recently, AIDE~\cite{su2024aide} mitigates novel object detection by retrieving and auto-labeling data from existing datasets. However, it is limited to detection models, and retrieval alone may lack data diversity.
Contemporary works~\cite{li2025avd2} train specialized transformers to analyze driving accident causes but do not use these insights to improve E2E models. In contrast, our CorrectAD identifies failure causes from E2E reasoning results, including perception, prediction, and planning. This enables data generation tailored to these failure points, enhancing model diversity and effectiveness. In addition, through fully automated iterative cycles, CorrectAD can continuously optimize performance.

\mypara{End-to-end Autonomous Driving.}
E2E models have garnered significant attention in autonomous driving by integrating perception, prediction, decision-making, and planning into a single framework~\cite{hu2023uniad,chen2024vadv2,cui2025drivemlm}. 
STP3~\cite{hu2022stp3} employs spatiotemporal feature learning to boost perception, prediction, and planning. UniAD~\cite{hu2023uniad} combines multiple perception and prediction tasks to improve planning. VAD\cite{Jiang2023VADVS} leverages vectorized scene representation to streamline planning, eliminating the need for dense maps, while VADv2\cite{chen2024vadv2} uses probabilistic planning and multi-view image sequences to predict control actions. In this paper, we utilize the notable and open-sourced UniAD~\cite{hu2023uniad} and VAD~\cite{Jiang2023VADVS}, along with our in-house E2E model to verify the effectiveness of our CorrectAD framework.

\begin{figure*}[t!]
    \centering
      \includegraphics[width=1.0\linewidth]{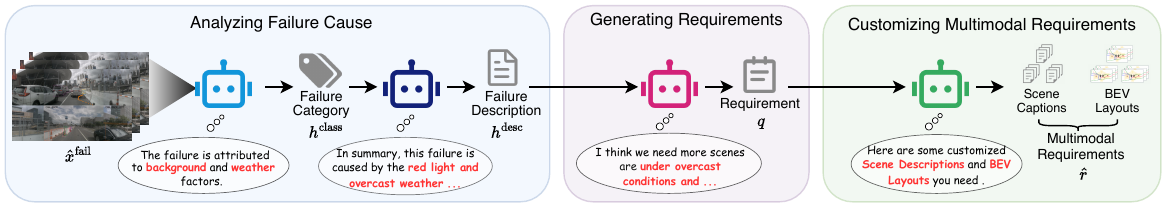}
    \caption{ \textbf{The framework of PM-Agent}. 
    Given a failure case $\hat{x}^{\text{fail}}$, PM-Agent first classifies the failure causes to $h^\text{class}$, then analyzes the failure description $h^\text{desc}$ in detail. Based on $h^\text{desc}$, PM-Agent generates specific requirements $q$. Then PM-Agent formulates multimodal requirements $\hat{r}$ (including bird's-eye-view layouts and scene captions) similar to the failure case to interact with the later generative model.
    }
    \label{fig: method_tailoragent} 
    \vspace{-0.2cm}
\end{figure*}

\mypara{Multi-view Video Generation.} Video generation is crucial for visual understanding. 
Recent advances in diffusion models for image generation~\cite{nichol2021_glide, rombach2022_ldm, ruiz2023dreambooth} have led to their use in video generation~\cite{harvey2022flexible, hoppe2022diffusion,ma2024unleashing_delphi,jiang2024dive,tang2025omnigen}, improving realism, control, and consistency. 
BEVGen~\citep{swerdlow2023_bevgen} first generates street images based on bird's-eye-view (BEV) layouts, while BEVControl~\cite{yang2023bevcontrol} creates foregrounds and backgrounds in two stages with a diffusion model. Magicdrive~\cite{gao2023magicdrive} applies ControlNet~\cite{zhang2023_controlnet} to inject BEV layouts. Later methods~\cite{wen2023panacea, wang2023drivedreamer, zhao2024drivedreamer-2} extend this for videos with cross-frame attention. Some works~\cite{wang2023drivedreamer, wen2023panacea, lu2025wovogen, xie2025glad, gao2024magicdrive-v2} introduce layout-conditioned video generation to diversify training data for perception models. GAIA-1~\cite{hu2023gaia} and ADriver-I~\cite{jia2023adriver} integrate LLMs for video generation, and DriveDreamer-2~\cite{zhao2024drivedreamer-2} uses a text-based traffic simulation for diverse driving videos. These methods face the challenge of low controllability and poor sequential consistency. It is worth noting that the Diffusion Transformer (DiT) paradigm, exemplified by Sora, has made remarkable progress in video generation. We improve spatiotemporal consistency by extending DiT to a multi-view setting in autonomous driving, which requires high-level geometric control, thus providing high-quality training data for E2E model.
\section{Method}

\subsection{Preliminary}
\label{sec:method_Preliminary}

\mypara{Definition of Failure Cases.}
Given a dataset \( D = \{ D^\text{train}, D^\text{val} \} \), \( D = ( X, Y) = \{ x_i, y_i \}_{i=1}^{|D|} \) consists of multi-view videos \( x_i = \{ x_i^j \}_{j=1}^{N_\text{view}} \) and corresponding 3D bboxes and map labels \( y_i \). A failure case occurs when, following the planned trajectory for the next \( T_{\text{e2e}} \) timesteps from the E2E model \( \mathcal{F} \), at least one collision occurs between the ego and others \( V_{\text{other}} = \{ v_j \}_{j=1}^{|V_{\text{other}}|} \) (including vehicles, pedestrian and barriers). Formally, the failure cases are defined as:
\begin{equation}
    \begin{aligned}
    D^{\text{fail}} = & \{ (X, Y) \in D^{\text{train}} \mid \exists t \leq T_{\text{e2e}}, \exists j \leq |V_{\text{other}}|, \\
    & \| \mathbf{p}_{\text{ego}}(t) - \mathbf{p}_{\text{other}}^j(t) \| < \epsilon  \},
    \end{aligned}
\end{equation}
where \( \mathbf{p}(t) \) is the vehicle's position at time $t$, \( \|\cdot\| \) is the euclidean distance, and \( \epsilon \) is the safety threshold.

\mypara{Pre-identification of Failure Categories.}
To precisely analyze failures, we pre-identify the categories of failure causes in $D$.
We use expert-annotated (details see Appendix) descriptions of failure causes  $Y^\text{desc} = \{ y_i^\text{desc}  \}_{i=1}^{N_\text{anno}}$ from  $N_\text{anno}$  failure cases. We use LLM to extract keywords  $Y^\text{key}$ and apply an adaptive clustering algorithm to obtain $K$ classes of causes $S = \{ S_k \}_{k=1}^K$. The process is denoted as:
\begin{equation}
y_i^\text{key} = \mathcal{LLM}(y_i^\text{desc})
\end{equation}
\begin{equation}
S_k = \{ y_i^\text{key} \in Y^\text{key} | {\textbf{d}}(y_i^\text{key}, s_k) \leq {\textbf{d}}(y_i^\text{key}, s_j), \forall j \neq k \},
\end{equation}
where  $s_k$ is the center of the  $k$-th cluster, and ${\textbf{d}}(\cdot, \cdot)$ is the two points' distance.
Then, we summarize the common cause features $l_k$ contained in each cluster $S_k$ for later CorrectAD, resulting in all possible failure categories $L = \{ l_k \}_{k=1}^K,  ~\text{where}~ l_k = \mathcal{LLM}(S_k)$.

\subsection{CorrectAD Overview}
The goal of CorrectAD is to generate new training data \( D^{gen} \) to specifically optimize failure cases \( D^\text{fail} \) of the E2E model \( \mathcal{F} \), producing an updated \( \mathcal{F'} \). 
At first, we preprocess the dataset: $ D \leftarrow (X', C, E ) = \{ (x'_i, c_i, e_i) \}_{i=1}^{|D|} $, where \( x'_i = \text{concat}(x_i) \) represents the operation of concatenating the multi-view videos \( x_i \) in a cyclic order into a single large video \( x'_i \), \( c_i = \mathcal{VLM}(x'_i) \) represents the scene caption of the video \( x'_i \), and \( e_i \leftarrow \text{project}(y_i) \) represents the BEV layout projected from BEV space into camera space. A similar definition applies to \( D^\text{train}, D^\text{val}\), and \(D^\text{fail} \).

To address the aforementioned challenge of generating new training data specifically for failure cases, we propose an automated data loop: First, the product manager, \textit{i.e.}, \textbf{PM-Agent} \( \mathcal{A} \), analyzes the failure and formulates multimodal requirements: 
$ R \leftarrow \mathcal{A}(D^\text{fail}). $
Next, the data department, \textit{i.e.}, \textbf{DriveSora} \( \mathcal{\textbf{G}} \), generates the new training data:
$ D^{gen} \leftarrow \{(X^{gen}, R) \mid X^{gen} = \mathcal{\textbf{G}}(R) \} $.
Then, \( \mathcal{F} \) is updated by fine-tuning it on both old and new training data, followed by evaluation on $D^\text{train}$ and iteration again.

\subsection{PM-Agent}
\label{sec:method_TailorAgent}

Since there is no effective way to link failure cases to the 3D generative model \(\mathcal{\textbf{G}}\), we propose the PM-Agent, as shown in Fig.~\ref{fig: method_tailoragent}, similar to a product manager, to bridge this gap by formulating 3D multimodal requirements \(R\).

\mypara{Analyzing Failure Cause.} It is essential for precisely customizing requirements. The vanilla baseline uses one-step VLMs conversation. But this yields suboptimal accuracy due to VLMs' limitation in reasoning over complex tasks. We propose a multi-round inquiry strategy to decompose the task: first, classifying the cause, then analyzing the failure in detail.
We first plot the output $o^\text{fail}$ from $\mathcal{F}$ onto failure cases, resulting $\hat{x}^\text{fail} = \text{plot}( {x'}^\text{fail}, o^\text{fail})$, where $o^\text{fail}$ includes detection, prediction and planning output for the next \( T_{\text{e2e}} \) timesteps.
Next, we guide the VLMs to classify the failure cause, outputting the failure category $h^\text{class}$:
\begin{equation}
\begin{aligned}
h^\text{class} = \mathcal{VLM}(\hat{x}^\text{fail}, L) = \{ l_i \in L \mid \text{\textbf{q}}(l_i \mid \hat{x}^\text{fail}) \ge \tau \},
\end{aligned}
\end{equation}
where \( \text{\textbf{q}}(\cdot \mid \cdot) \) is the probability that the later belongs to the former,  \( \tau \) is the classification threshold.
Based on the classification result, we then perform a specificly analysis of the failure cause description $h^\text{desc} $:
\begin{equation}
h^\text{desc} = \mathcal{VLM}(\hat{x}^\text{fail}, h^\text{class}).
\end{equation}
\mypara{Generating Requirements.}
These requirements are essential for understanding the context and the details surrounding the failure, which will guide $\mathcal{\textbf{G}}$ to generate the desired data. For each failure case, we generate a requirement \( q \) based on both the class $h^\text{class}$ and description $h^\text{desc}$ of the failure cause:
\begin{equation}
q = \mathcal{LLM}( h^\text{class}, h^\text{desc}).
\end{equation}
\mypara{Formulating Multimodal Requirements.}
To better interface with $\mathcal{\textbf{G}}$, we select the top-\( K \) samples from \( D^{\text{train}} \) whose scene captions \( c \) are most similar to \( q \) and extract the corresponding BEV layouts \( e \) to assemble the multimodal requirements \( \hat{r} \):
\begin{equation}
\hat{r} = \mathcal{VLM}(q, D^{\text{train}}) = \{ (c, e) \mid \textbf{s}(c, q) \geq \delta \},
\end{equation}
where $\textbf{s}(\cdot, \cdot)$ represents the similarity calculation, \( \delta \) is the similarity threshold. 
Finally, the union of all \( \hat{r}\), denoted as $R = \{ \hat{r}_i\}_{i=1}^{|R|}$, serves as the set of multimodal requirements for the current iteration.

\subsection{DriveSora}
\label{sec:method_DriveSora}

\begin{figure}[t]
    \centering
    \vspace{-0.3cm}
      \includegraphics[width=0.95\linewidth]{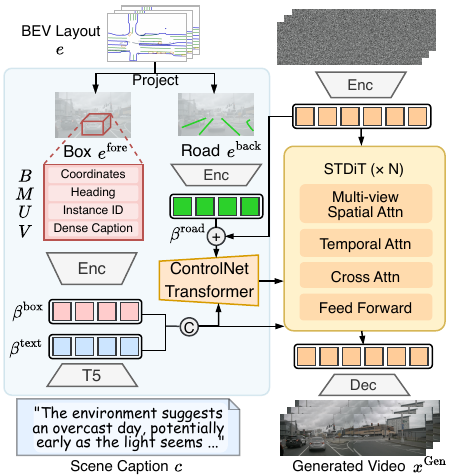}
    \caption{
    \textbf{\textbf{The framework of DriveSora},} which performs data generation tasks, aiming to produce high-quality, diverse new data.
    }
    \label{fig: method_drivesora} 
    \vspace{-0.1cm}
\end{figure}

Since previous generative works struggle with the quality of generated data, we propose DriveSora $\mathcal{\textbf{G}}$, akin to a data department, by specifically generating high-fidelity training data $D^{\text{gen}}$ to enhance the ability of the E2E model $\mathcal{F}$ against complex scenario. As shown in Fig.~\ref{fig: method_drivesora}, DriveSora takes the multimodal prompt $R$ as input, based on the Spatial-Temporal Diffusion Transformer (STDiT) architecture to generate videos $X^{\text{gen}}=\{ x^\text{gen}_i\}_{i=1}^{|X^{\text{gen}}|}$, where $ x^\text{gen}_i$ represents generated video which consists of \( T_\text{frame} \) frames and \( N_\text{view} \) views.

\mypara{Multimodal Control Generation.}
We first improve generation fidelity by encoding more fine-grained conditions. The input multimodal prompt includes the scene caption \( c \) and the BEV layout \( e \), where \( e \) is first decoupled into the foreground layout \( e^\text{fore} \) and the background layout \( e^\text{back} \). \( e^\text{fore} = (B, M, U, V) = \{(b_\text{n}, m_\text{n}, u_\text{n}, v_\text{n})\}_{n=1}^{|N_\text{view}|}\), where $b_\text{n}\in{[0,1]}^{N^\text{box} \times 4}$ means bbox coordinates, $m_\text{n}\in{[-180, 180)}^{N^\text{box} \times 1}$ means heading, $u_\text{n}\in{[0,1]}^{N^\text{box} \times 1}$ means instance id, $v_\text{n}\in \mathbb{R}^{N^\text{box} \times 1}$ means dense caption, and $N^\text{box}$ means the number of boxes. $e^\text{back} \in \mathbb{R}^{ H \times W \times 3 }$ means colored lines for road maps. To obtain the box embedding $\beta^\text{box}$, road embedding $\beta^\text{road}$ and text embedding $\beta^\text{text}$, the encoding process is:
\begin{equation}
    \begin{aligned}
        & {\beta^\text{box} = \textbf{Mlp}}( \textbf{Fe}({B}) +  \textbf{Fe}({M}) + \textbf{Fe}({\text{U}}) + \textbf{E}_{\textbf{text}}({\text{V}}) ), \\[1mm]
        & ~~~~~ {\beta^\text{road}} = \textbf{E}_{\textbf{image}}(\alpha), 
        ~~~~~~ {\beta^\text{text}} = \textbf{E}_{\textbf{text}}(c),
    \end{aligned}
\label{equ:multimodal_encoding}
\end{equation}
where $\textbf{Fe}(\cdot)$ is the Fourier Embedder~\cite{mildenhall2021nerf}, $\textbf{E}_{\textbf{text}}$ is the T5 Encoder~\cite{raffel2020exploring_t5}, and $\textbf{E}_{\textbf{image}}$ is the VAE~\cite{rombach2022_ldm}. 
We concatenate box embedding $\beta^\text{box}$ and text embedding $\beta^\text{text}$ to enable text and vehicle control through cross-attention (\textbf{CA}) in STDiT:
\begin{equation}
    \begin{aligned}
    {q} = \text{\textbf{Lin}}({z_{in}}),~~
    {k} &= \text{\textbf{Lin}}([\beta^\text{box}, \beta^\text{text}]), ~~
    {v} = \text{\textbf{Lin}}([\beta^\text{box}, \beta^\text{text}]), \\
     \text{\textbf{CA}}(q, k, v) &= \text{\textbf{Softmax}}(\frac{q \cdot k^{T}}{\sqrt{d}})\cdot v,
    \end{aligned}
\label{equ:text_cross_attention}
\end{equation}
where $\text{\textbf{Lin}}(\cdot)$ is a linear layer, and $z_{in} \sim \mathcal{N}(0,1)$ is the noise latents. Following ControlNet~\cite{zhang2023_controlnet}, we add a trainable ControlNet-Transformer to STDiT for precise layout control with road embedding $\beta^\text{road}$. The STDiT block's calculation process is formulated as:
\begin{equation}
    \begin{aligned}
        z_{out} = \text{\textbf{STDiT}}(z_{in}) + \text{\textbf{Zero}}(\text{\textbf{Control}}(z_{in} + \beta^\text{road})),
    \end{aligned}
\label{equ:STDiT}
\end{equation}
where $\text{\textbf{Zero}}(\cdot)$ is zero-initialized trainable convolution layers, and $\text{\textbf{Control}}(\cdot)$ is the ControlNet-Transformer, which is detailed in Appendix.

\mypara{Parameter-free Multi-view Spatial Attention.}
To enhance spatial consistency, we extend STDiT's Self-Attention with Multi-View Self-Attention (MVA). Unlike prior works using additional cross-view attention~\cite{gao2023magicdrive, wen2023panacea}, our parameter-free approach reshapes $z_{in} \in \mathbb{R}^{(BV)~\times~(TS)~\times~C}$ to $z_{in}'\in \mathbb{R}^{(BT)~\times~(VS)~\times~C}$ ($S$ is embedding resolution) and applies self-attention directly:
\begin{equation}
    \begin{aligned}
    {z_{in}'} =& ~~\text{\textbf{Reshape}}({z_{in}}), \\[1.95mm]
    {q} = \text{\textbf{Lin}}({z_{in}'}),~~
    {k} &= \text{\textbf{Lin}}({z_{in}'}),~~
    {v} = \text{\textbf{Lin}}({z_{in}'}), \\
     \text{\textbf{MVA}}(q, k, v) &= \text{\textbf{Softmax}}(\frac{q \cdot k^{T}}{\sqrt{d}})\cdot v.
    \end{aligned}
\label{equ:Multi_view_Spatial_Attention}
\end{equation}

\mypara{Multi-conditional Classifier-free Guidance.}
We improve the condition-content alignment by conditional and unconditional denoising mode. Unlike~\cite{gao2023magicdrive}, which concurrently sets all conditions to null $\phi$ in the unconditional mode, we alternately nullify each condition to strengthen individual guidance. 
The generator \( \mathcal{\textbf{G}}_{\theta} ( z_\text{in}, e^\text{fore}, e^\text{back}, c) \) takes box, road, and text conditions with guidance scales $\lambda_\text{fore}, \lambda_\text{back}, \lambda_\text{text}$. During training, we set each condition to $\phi$ independently with a 5\% probability, 
and all jointly with the same rate.
During inference, the process is formulated as:
\begin{equation}
    \begin{aligned}
        \tilde{\mathcal{\textbf{G}}}_{\theta}( &z_\text{in} , e^\text{fore} , c_ R, c) = \mathcal{\textbf{G}}_{\theta}(z_\text{in}, \phi, \phi, \phi) \\
        &+\lambda_\text{text}\cdot(\mathcal{\textbf{G}}_{\theta}(z_\text{in}, \phi, \phi, c)-\mathcal{\textbf{G}}_{\theta}(z_\text{in}, \phi,\phi,\phi)) \\
        &+\lambda_\text{back}\cdot(\mathcal{\textbf{G}}_{\theta}(z_\text{in}, \phi, e^\text{back}, c)-\mathcal{\textbf{G}}_{\theta}(z_\text{in}, \phi, \phi, c)) \\
        &+\lambda_\text{fore}\cdot(\mathcal{\textbf{G}}_{\theta}(z_\text{in}, e^\text{fore}, e^\text{back}, c)-\mathcal{\textbf{G}}_{\theta}(z_\text{in}, \phi, e^\text{back}, c)).
    \end{aligned}
\label{equ:cfg}
\end{equation}
\section{Experiments}
\label{experiments}

\setlength{\tabcolsep}{3.5pt}
\begin{table}[t!]
\centering
\small
\begin{tabular}{
>{\columncolor[HTML]{FFFFFF}}l |
>{\columncolor[HTML]{FFFFFF}}c 
>{\columncolor[HTML]{FFFFFF}}c 
>{\columncolor[HTML]{FFFFFF}}c 
>{\columncolor[HTML]{FFFFFF}}c |
>{\columncolor[HTML]{FFFFFF}}c 
>{\columncolor[HTML]{FFFFFF}}c 
>{\columncolor[HTML]{FFFFFF}}c 
>{\columncolor[HTML]{FFFFFF}}c }
\toprule
\multicolumn{1}{c|}{\cellcolor[HTML]{FFFFFF}}                                    & \multicolumn{4}{c|}{\textbf{L2 (m) $\downarrow$}}   & \multicolumn{4}{c}{\textbf{Collision (\%) $\downarrow$}}     \\
\multicolumn{1}{l|}{\multirow{-2}{*}{\cellcolor[HTML]{FFFFFF}\textbf{Method}}} & \textbf{1s}   & \textbf{2s}    & \textbf{3s}   & \textbf{Avg.} & \textbf{1s}   & \textbf{2s}            & \textbf{3s}   & \multicolumn{1}{l}{\cellcolor[HTML]{FFFFFF}\textbf{Avg.}} \\
\midrule
\multicolumn{9}{l}{\cellcolor[HTML]{FFFFFF}\textit{\textbf{UniAD metrics}}}        \\
\midrule

NMP
& -             & -             & 2.31          & -             & -             & -             & 1.92          & -                                                \\
SA-NMP
& -             & -             & 2.05          & -             & -             & -             & 1.59          & -                                                \\
FF
& 0.55          & 1.20           & 2.54          & 1.43          & 0.06          & 0.17          & 1.07          & 0.43                                             \\
EO
& 0.67          & 1.36          & 2.78          & 1.60           & 0.04          & 0.09          & 0.88          & 0.33                                             \\
UniAD
& \textbf{0.48}          & 0.96          & 1.65          & 1.03          & 0.05          & 0.17          & 0.71          & 0.31                                             \\
AIDE$^{\ast}$
& 0.51          & 0.96          & 1.60          & 1.02          & 0.05          & 0.16          & 0.64          & 0.28                                             \\
\rowcolor{gray!15} 
\textbf{CorrectAD$^{\ast}$}                         &0.50 	&\textbf{0.92} 	&\textbf{1.53} 	&\textbf{0.98}  & \textbf{0.02} & \textbf{0.14} & \textbf{0.42} & \textbf{0.19}                                    \\
\midrule
\multicolumn{9}{l}{\cellcolor[HTML]{FFFFFF}\textit{\textbf{ST-P3 Metrics}}}                                                                                                                                                  \\
\midrule
ST-P3
& 1.33          & 2.11          & 2.90           & 2.11          & 0.23          & 0.62          & 1.27          & 0.71                                             \\
VAD
& 0.41          & 0.70           & 1.05          & 0.72          & 0.07          & 0.17          & 0.41          & 0.22                                             \\
AIDE$^{\dag}$ 
& 0.39	& 0.68	& 1.01	& 0.69	& 0.06	& 0.17	& 0.42	& 0.22 \\
\rowcolor{gray!15} 
\textbf{CorrectAD$^{\dag}$ }                         & \textbf{0.34} & \textbf{0.60} & \textbf{0.94} & \textbf{0.62} & \textbf{0.05} & \textbf{0.14} & \textbf{0.40} &  \textbf{0.20}       \\
\bottomrule
\end{tabular}
\caption{
E2E planning comparison on nuScenes validation set. $^{\ast}$ and $^{\dag}$ denotes frameworks initialized by UniAD
and VAD, respectively.
}
\label{tab:SeCoDE_framework}
\end{table}

\setlength{\tabcolsep}{3.8pt}
\begin{table}[t!]
\centering
\small
\begin{tabular}{l|cccc|cccc}
\toprule
 & \multicolumn{4}{c|}{\textbf{L2 (m) $\downarrow$}}                                                                  & \multicolumn{4}{c}{\textbf{Hit Rate (\%) $\uparrow$}}       \\
\multirow{-2}{*}{\textbf{Method}} & \multicolumn{1}{l}{\textbf{1s}}       & \multicolumn{1}{l}{\textbf{3s}}       & \multicolumn{1}{l}{\textbf{8s}}       & \multicolumn{1}{l|}{\textbf{Avg.}} & \multicolumn{1}{l}{\textbf{1s}} & \multicolumn{1}{l}{\textbf{3s}}       & \multicolumn{1}{l}{\textbf{8s}}       & \multicolumn{1}{l}{\textbf{Avg.}} \\
\midrule
Baseline   & 0.10   & 0.54   & 1.91   & 0.85       & 0.98   & 0.80   & 0.53   & 0.77  \\
\midrule
AIDE$^{\ddag}$
& 0.09	&0.50	&1.79	&0.79	&0.98	&0.81	&0.54	&0.78 \\
\rowcolor{gray!15} 
\textbf{CorrectAD$^{\ddag}$}         & \textbf{0.08} & \textbf{0.44} & \textbf{1.33} & \textbf{0.62}                     & \textbf{0.99}                   & \textbf{0.83} & \textbf{0.63} & \textbf{0.82}   \\
\bottomrule
\end{tabular}
\caption{E2E planning comparison on a large in-house validation set. ``Hit Rate" indicates the recall rate of the planned trajectory relative to the real trajectory at different timesteps. $^{\ddag}$ denotes framework initialized by Baseline (our in-house E2E model).
}
\label{tab:SeCoDE_OnInhouseDataset}
\end{table}

\subsection{Experimental Setting}

\mypara{Dataset.}
We evaluate on two datasets: (1) the real-world nuScenes~\cite{caesar2020_nuscenes} dataset with 700 training and 150 validation scenes of 20s 6-view videos at 12Hz;
(2) a more challenging in-house E2E dataset with diverse driving behaviors, containing 3M training and 0.6M validation scenes of 15s 6-view videos at 10Hz. Behavior distribution is detailed in the Appendix.

\mypara{Metrics.}
We evaluate CorrectAD in three E2E models: UniAD~\cite{hu2023uniad}, VAD~\cite{Jiang2023VADVS} (using L2 error and collision rate), and our in-house E2E model (using L2 error and hit rate). 
For PM-Agent, we assess its analysis ability using the accuracy of the failure category and the semantic distance of the descriptions. For DriveSora, we assess the fidelity and consistency of the generated videos
(using FID~\cite{heusel2017_fid}, FVD~\cite{unterthiner2018fvd}, and CLIP score~\cite{yang2023bevcontrol}), and detection score (using NDS~\cite{wang2023streampetr}) to measure the sim-to-real gap.

\mypara{Methods for Comparison.} 
To our knowledge, little work focuses on automated data pipeline for self-correcting failures in autonomous driving E2E models, making it difficult to find a fully comparable counterpart for CorrectAD. However, we noticed AIDE~\cite{su2024aide}, a closed-source method for novel object detection tasks, which shares a similar process: identifying issues, curating data, updating the model, and verifying results. Key differences include: \textbf{1)} AIDE targets detection tasks, while our method focuses on E2E planning tasks; \textbf{2)} AIDE retrieves data from existing dataset, while we generate new data using a generative model. To ensure a fair comparison, we re-implemented AIDE’s process for the planning task in this paper. Details are in the Appendix.

\subsection{Main Results}

\begin{figure}[t]
    \centering
    \vspace{-0.1cm}
    \includegraphics[width=\linewidth]{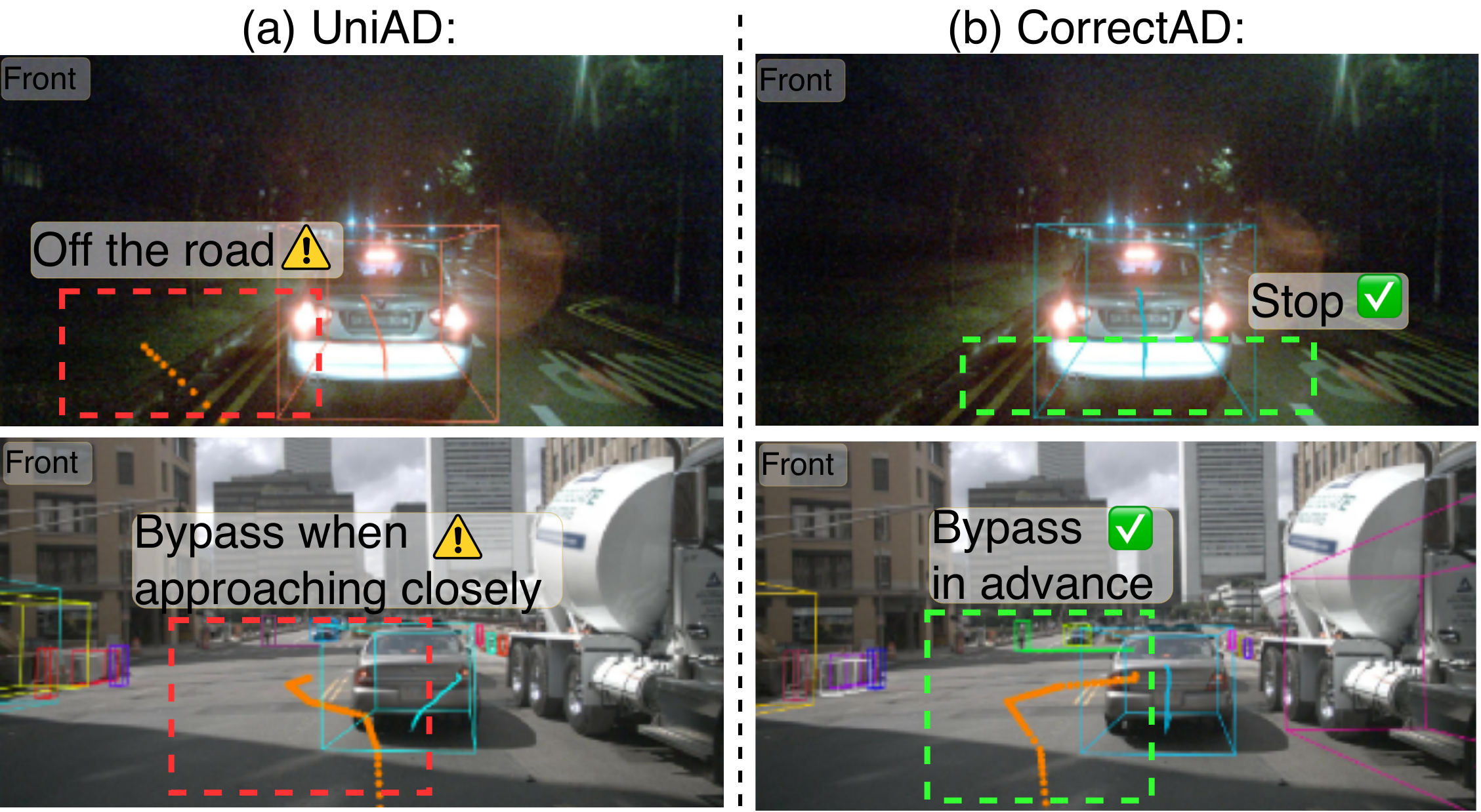}
    \caption{
    Visualization of two examples before and after \textbf{self-correction on nuScenes validation set.} \textbf{(a)} We show two hard examples from the validation set, ``a low-visibility night'', ``bypass in dense traffic flow''. \textbf{(b)} Our framework can fix these examples. 
    }
    \label{fig:uniad generalization comparison} 
\end{figure}

\begin{figure}[t]
    \centering
      \includegraphics[width=\linewidth]{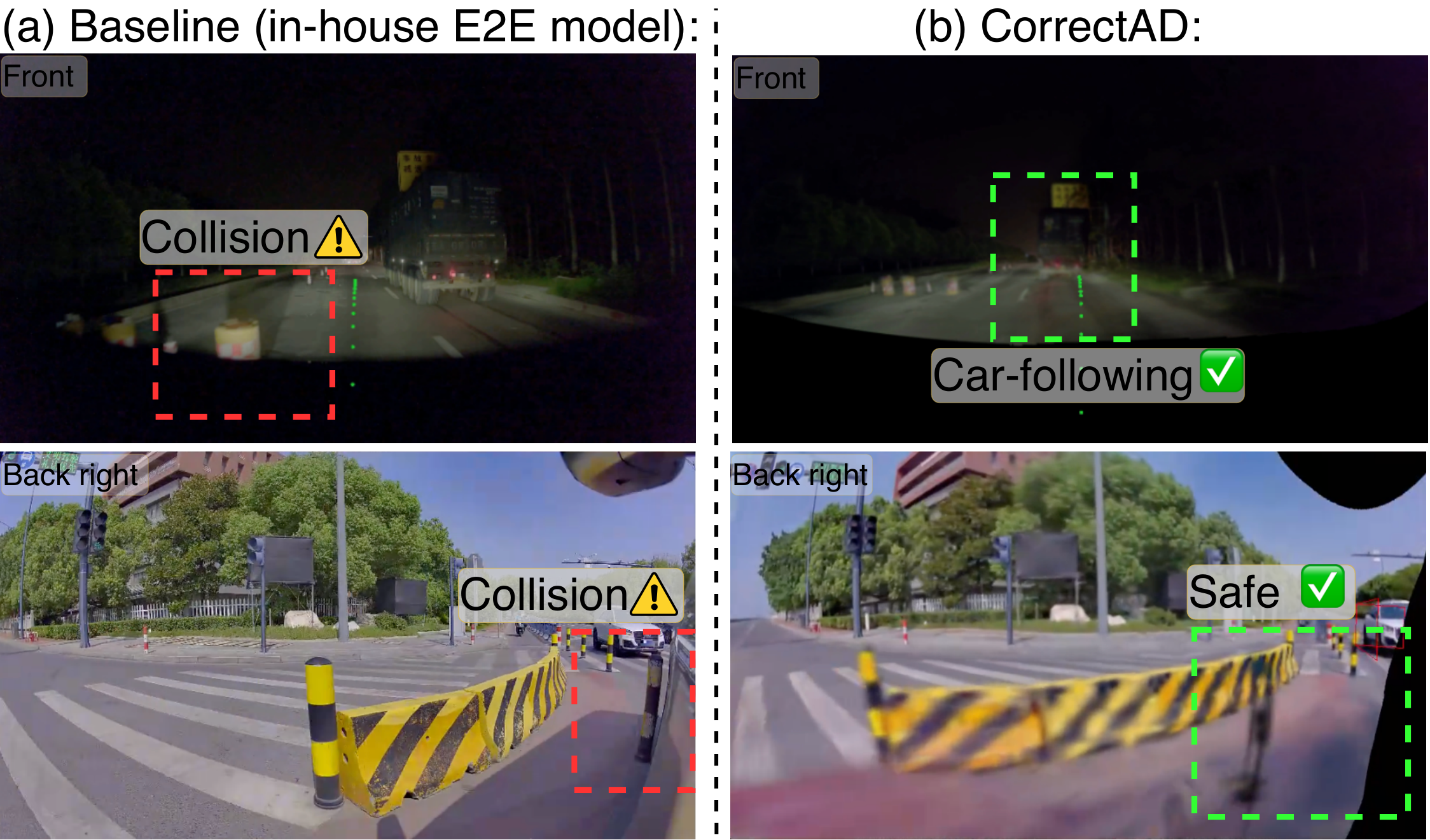}
    \caption{Visualization of two examples before and after \textbf{self-correction on our in-house validation set.} Results are rendered via a proprietary closed-loop simulator based on Gaussian splatting.}
    \label{fig:liauto_self_corrections} 
\end{figure}

Evaluating CorrectAD against state-of-the-art methods on the nuScenes validation set, our framework achieves superior performance in both L2 and collision rate metrics (see Tab.~\ref{tab:SeCoDE_framework}). In contrast to AIDE, which only retrieves training data, CorrectAD improves safety metrics by analyzing failure causes and specifically generating new training data. We also show how our CorrectAD achieves self-correction in Fig.~\ref{fig:uniad generalization comparison}. Only the front view is shown here for clarity. All multi-view results are in the Appendix.

Furthermore, evaluating on the large in-house E2E model (see Tab.~\ref{tab:SeCoDE_OnInhouseDataset}), CorrectAD significantly outperforms AIDE in L2 error and hit rate, demonstrating strong generalization capability across different E2E models. Fig.~\ref{fig:liauto_self_corrections} shows the self-correction results on a large in-house dataset, which is visualized via our proprietary closed-loop simulator based on Gaussian Splatting~\cite{yan2024street}, demonstrating effectiveness in fixing failures.

\mypara{Statistical distribution of augmented data.}
To better understand why our method significantly outperforms the AIDE baseline in enhancing the performance of the E2E model, we visualize the statistical distribution of the augmented data each method provides (see Fig.~\ref{fig:gendata_distribution_gap}). A detailed explanation of our visualization approach is available in the Appendix.
The rightmost column in the figure highlights the distribution of failure cases in the validation set, arguably the most critical distribution for the E2E planning model to learn from. Notably, the data generated by our method exhibit a much closer alignment with this failure distribution compared to other methods. This strong alignment is a key factor that enables our approach to deliver superior effectiveness.

\subsection{Ablation Studies}

\setlength{\tabcolsep}{4.2pt}
\begin{table}[t!]
\centering
\small
\begin{tabular}{cc|cccc|cccc}
\toprule
 &  & \multicolumn{4}{c|}{\textbf{L2 (m) $\downarrow$}} & \multicolumn{4}{c}{\textbf{Collision (\%) $\downarrow$}} \\
\multirow{-2}{*}{\textbf{(1)}} & \multirow{-2}{*}{\textbf{(2)}} & \textbf{1s} & \textbf{2s} & \textbf{3s} & \textbf{Avg.} & \textbf{1s} & \textbf{2s} & \textbf{3s} & \multicolumn{1}{l}{\textbf{Avg.}} \\
\midrule
{\color[HTML]{C0C0C0} \XSolidBrush} & {\color[HTML]{C0C0C0} \XSolidBrush} & 0.54 & 1.03 & 1.71 & 1.09 & 0.05 & 0.18 & 0.81 & 0.35 \\
{\color[HTML]{C0C0C0} \XSolidBrush} & \Checkmark & 0.53 & 0.99 & 1.66 & 1.06 & 0.10 & 0.20 & 0.62 & 0.31 \\
\Checkmark & {\color[HTML]{C0C0C0} \XSolidBrush} & 0.52 & 0.96 & 1.62 & 1.03 & 0.08 & 0.20 & 0.58 & 0.29 \\
\rowcolor{gray!15} 
\Checkmark & \Checkmark &\textbf{0.50} 	&\textbf{0.92} 	&\textbf{1.53} 	&\textbf{0.98}  & \textbf{0.02} & \textbf{0.14} & \textbf{0.42} & \textbf{0.19} \\
\bottomrule
\end{tabular}
\caption{Ablation on (1) PM-Agent and (2) DriveSora.}
\label{tab:Ablation_TailorAgent&Delphi}
\vspace{-0.1in}
\end{table}

\setlength{\tabcolsep}{3.0pt}
\begin{table}[t!]
\centering
\small
\begin{tabular}{l|cccc}
\toprule
Method & \begin{tabular}[c]{@{}c@{}}Foreground\\ acc.↑\end{tabular} & \begin{tabular}[c]{@{}c@{}}Background\\ acc.↑\end{tabular} & \begin{tabular}[c]{@{}c@{}}Weather\\ acc.↑\end{tabular} & \begin{tabular}[c]{@{}c@{}}Semantic\\ dist.↓\end{tabular} \\
\midrule
Baseline(1 step)
& {\color[HTML]{C0C0C0} N/A} & {\color[HTML]{C0C0C0} N/A} & {\color[HTML]{C0C0C0} N/A} & 4.72 \\
\rowcolor{gray!15} 
\textbf{PM-Agent}
& \textbf{92.59\%} & \textbf{87.41\%} & \textbf{91.85\%} & \textbf{3.49} \\
\bottomrule
\end{tabular}
\caption{Accuracy of VLM in analyzing failure causes. 
}
\label{tab:cause_acc}
\end{table}

\begin{figure}[t!]
    \centering
    \includegraphics[width=0.999\linewidth]{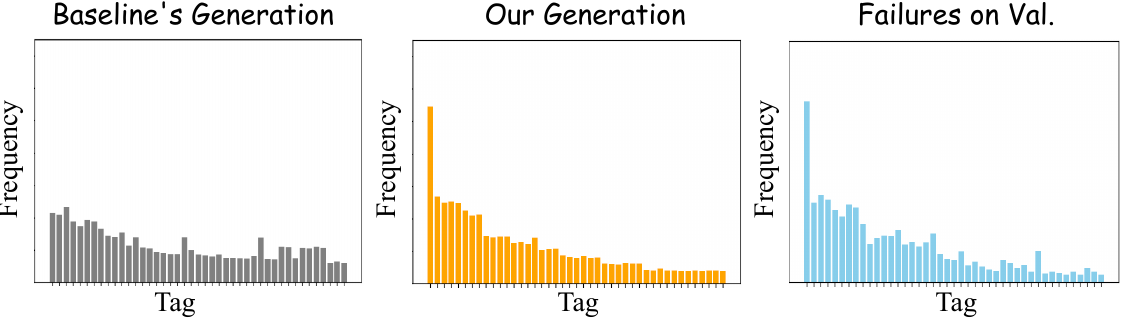}
    \caption{
    \textbf{Distribution gap} between generated data from AIDE baseline,
    our method, and failures on the validation set.
    }
    \label{fig:gendata_distribution_gap}
\end{figure}

\mypara{Ablation on proposed PM-Agent and DriveSora.}
To assess the individual contributions of the two proposed modules, we disable each in turn. 
In the first row of Tab.~\ref{tab:Ablation_TailorAgent&Delphi}, we use augmented data created by randomly duplicating samples from the training set. This yields no gain due to redundant data without meaningful distributional alignment.
Introducing DriveSora in the second row generates more diverse data, which partially mitigates this issue and leads to moderate improvements. As shown in the last two rows, incorporating PM-Agent to tailor the augmented data distribution to failure cases yields further gains.
Combining both DriveSora and PM-Agent, our full method achieves the best results: 0.98 L2 error and 0.19 collision rate, demonstrating the impact of using DriveSora for data diversity and PM-Agent for failure-focused distribution control. This validates the importance of both the distribution and diversity of the augmented data.

\mypara{The accuracy of PM-Agent.} 
Tab.~\ref{tab:cause_acc} compares PM-Agent's results with those obtained from a single direct prompt (one-step) to the VLM, where {\color[HTML]{C0C0C0} N/A} means not available due to baseline skipping analysis failure category. Specifically, we used the expert-annotated data, 
as the ground truth (GT).

\begin{figure}[t!]
        \centering
        \vspace{-0.3cm}
    \centering
    \includegraphics[width=0.99\linewidth]{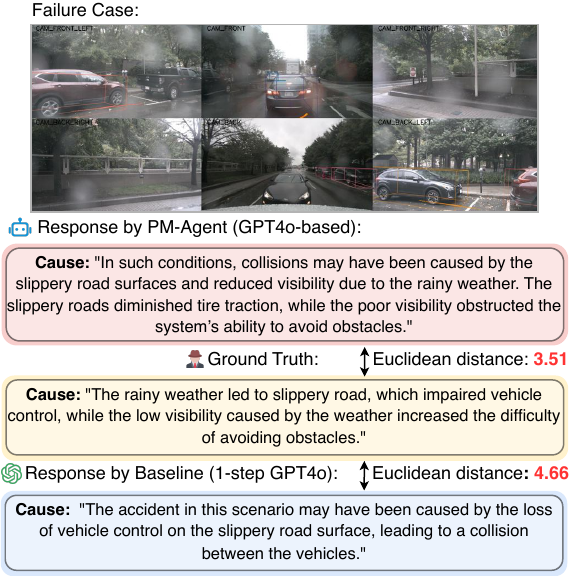}
    \caption{\textbf{An cause example} of GT and response by PM-Agent and baseline (one-step GPT4o).}
    \label{fig:agent_acc}
\end{figure}

Subsequently, we measured the degree of alignment between the different outputs and the GT by calculating the textual semantic distance. The VLM we chose is GPT-4o, and the results show that our PM-Agent is effective. We can find that, by decomposing complex tasks into a series of subtasks for multi-step reasoning, PM-Agent significantly improved accuracy, reducing the semantic distance from 4.72 to 3.49. As a reference, we provide visual cases scoring both 3.51 and 4.66 in Fig.~\ref{fig:agent_acc}. We emphasize that using VLM to analyze causes is an exploratory area in the field. Real-world failures are more complex, and we expect that the proposed paradigm can offer insights to the industry.

\setlength{\tabcolsep}{8.0pt}
\begin{table}[t]
\centering
\small
\begin{tabular}{l|cccc}
\toprule
\textbf{Generator} & {\textbf{FID$\downarrow$}} & {\textbf{CLIP$\uparrow$}}  & {\textbf{FVD$\downarrow$}} & {\textbf{NDS$\uparrow$}} \\
\midrule
BEVGen
& 25.54                     & 71.23                       & -              & {\color[HTML]{C0C0C0} N/A} \\
BEVControl
& 24.85                    & 82.70                      & -              & {\color[HTML]{C0C0C0} N/A} \\
DriveDreamer
& 26.8                     & {\color[HTML]{C0C0C0} N/A}                        & 353.2             & {\color[HTML]{C0C0C0} N/A} \\
DriveDreamer-2
& 25.0                     & {\color[HTML]{C0C0C0} N/A}                        & 105.1             & {\color[HTML]{C0C0C0} N/A} \\
WoVoGen
& 27.6 & {\color[HTML]{C0C0C0} N/A} & 417.7& {\color[HTML]{C0C0C0} N/A} \\
MagicDrive
& 16.20                    & 82.47                        & 221.90            & 34.56 \\
Panacea
& 16.96                    & 84.23                        & 139.0             & 32.10 \\
Drive-WM
& 15.80                    & {\color[HTML]{C0C0C0} N/A}                        & 122.7             & {\color[HTML]{C0C0C0} N/A} \\ 
MagicDrive-v2
& 20.91 & 85.25 & 94.84 & 35.79 \\
\rowcolor{gray!15} 
\textbf{DriveSora (Ours)}  & \textbf{15.08}           & \textbf{86.73}             & \textbf{94.51}    & \textbf{36.58} \\
\bottomrule
\end{tabular}
\caption{
Comparison of DriveSora with state-of-the-art generators in terms of consistency and controllability on the nuScenes validation set.
{\color[HTML]{C0C0C0} N/A} means not available due to closed-source.}
\label{tab:control}
\end{table}

\setlength{\tabcolsep}{3.1pt}
\begin{table}[t]
\centering
\small
\begin{tabular}{l|cccc|cccc}
\toprule
\multirow{2}{*}{\textbf{Generator}} & \multicolumn{4}{c|}{\textbf{L2 (m) $\downarrow$}} & \multicolumn{4}{c}{\textbf{Collision (\%) $\downarrow$}} \\
 & \textbf{1s} & \textbf{2s} & \textbf{3s} & \textbf{Avg.} & \textbf{1s} & \textbf{2s} & \textbf{3s} & \multicolumn{1}{l}{\textbf{Avg.}} \\
 \midrule
Panacea & \textbf{0.49} & 0.98 & 1.62 & 1.03 & 0.08 & 0.18 & 0.56 & 0.27 \\
MagicDrive-v2 & 0.50	& 0.96	& 1.55	& 1.00	& 0.05	& \textbf{0.13}	& 0.51	& 0.23 \\
\rowcolor{gray!15} 
 \textbf{DriveSora} &0.50 	&\textbf{0.92} 	&\textbf{1.53} 	&\textbf{0.98}  & \textbf{0.02} & 0.14 & \textbf{0.42} & \textbf{0.19} \\
\bottomrule
\end{tabular}
\caption{The effect of using different video generators in CorrectAD.}
\label{tab:Ablation_generator}
\end{table}

\begin{figure}[t!]
    \centering
    \vspace{-0.1in} 
    \includegraphics[width=0.999\linewidth]{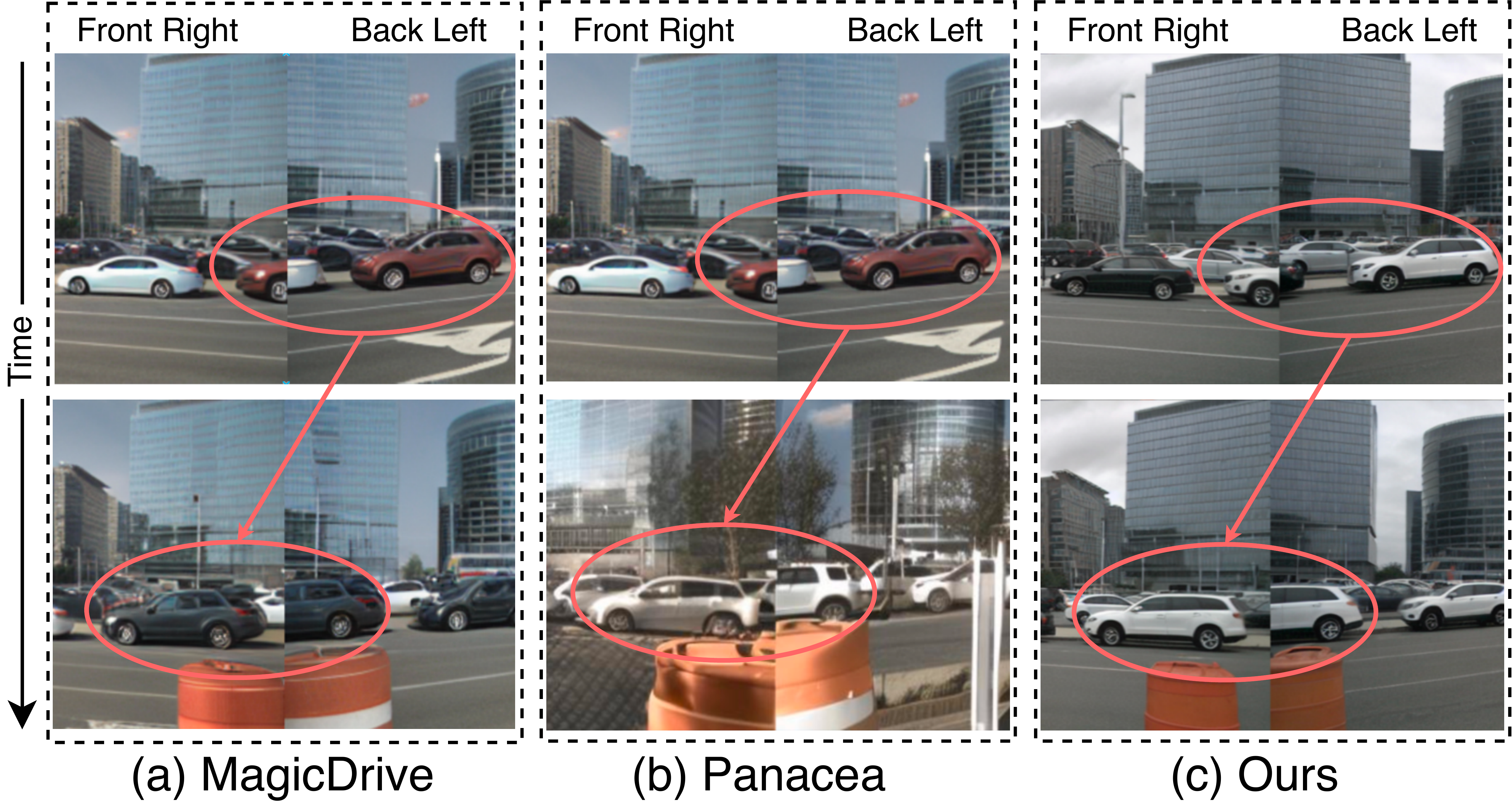}
    \caption{
    The visualization comparison of cross-frame consistency.
    }
    \vspace{-0.1in}
    \label{fig: visual_comparison_of_local_region} 
\end{figure}

\mypara{Comparison of the data quality generated by DriveSora.}
We assess the quality of video generation through a comprehensive evaluation including both quantitative and qualitative aspects, comparing our proposed DriveSora with previous generative methods.
In Tab.~\ref{tab:control}, we report metrics for three aspects: spatial and temporal consistency, and sim2real gap, on the nuScenes validation set. In short, our method surpasses state-of-the-art by a clear margin in video generation tasks. 
In Fig.~\ref{fig: visual_comparison_of_local_region}, we present videos generated by different methods on the same clip. Our method maintains a consistent spatial and temporal appearance, whereas the previous methods failed. 
It can be seen that our method has the powerful ability to generate high-quality videos with spatiotemporal consistency, which is beneficial for the training of E2E models.

\mypara{Effects using different video generators in CorrectAD.}
To further validate the impact of generated data quality on the performance of the E2E model, we replace the generative model within CorrectAD with an open source state-of-the-art method, Panacea. As illastrated in Tab.~\ref{tab:Ablation_generator}, the model trained with data generated by Panacea performs worse than the model trained with data from DriveSora, which highlights the importance of high-quality generated data for training E2E models.

\setlength{\tabcolsep}{3.5pt}
\begin{table}[t!]
\centering
\small
\begin{tabular}{l|c|cccc|cccc}
\toprule
\multirow{2}{*}{\textbf{Iter.}} & \multirow{2}{*}{\textbf{\begin{tabular}[c]{@{}c@{}}D-D $\downarrow$\end{tabular}}} & \multicolumn{4}{c|}{\textbf{L2 (m) $\downarrow$}} & \multicolumn{4}{c}{\textbf{Collision (\%) $\downarrow$}} \\
 &  & \textbf{1s} & \textbf{2s} & \textbf{3s} & \textbf{Avg.} & \textbf{1s} & \textbf{2s} & \textbf{3s} & \multicolumn{1}{l}{\textbf{Avg.}} \\
 \midrule
1 & 0.15 & \textbf{0.50} & 0.99 & 1.68 & 1.06 & 0.07 & 0.19 & 0.53 & 0.26 \\
2 & 0.11 & 0.51 & 0.96 & 1.65 & 1.04 & 0.04 & 0.17 & 0.46 & 0.22 \\
\rowcolor{gray!15} 
3 & \textbf{0.09} &\textbf{0.50} 	&\textbf{0.92} 	&\textbf{1.53} 	&\textbf{0.98}  & \textbf{0.02} & \textbf{0.14} & \textbf{0.42} & \textbf{0.19} \\
\bottomrule
\end{tabular}
\caption{The effect of multiple iterations of CorrectAD. ``Iter.'' means the number of iterations. The D-D metric represents the distribution of Hellinger Distance between the generated data and the failures in the validation set.}
\label{tab:Ablation_multi_iteration}
\end{table}

\begin{figure}[t]
    \centering
    \vspace{-0.1cm}
    \includegraphics[width=1.0\linewidth]{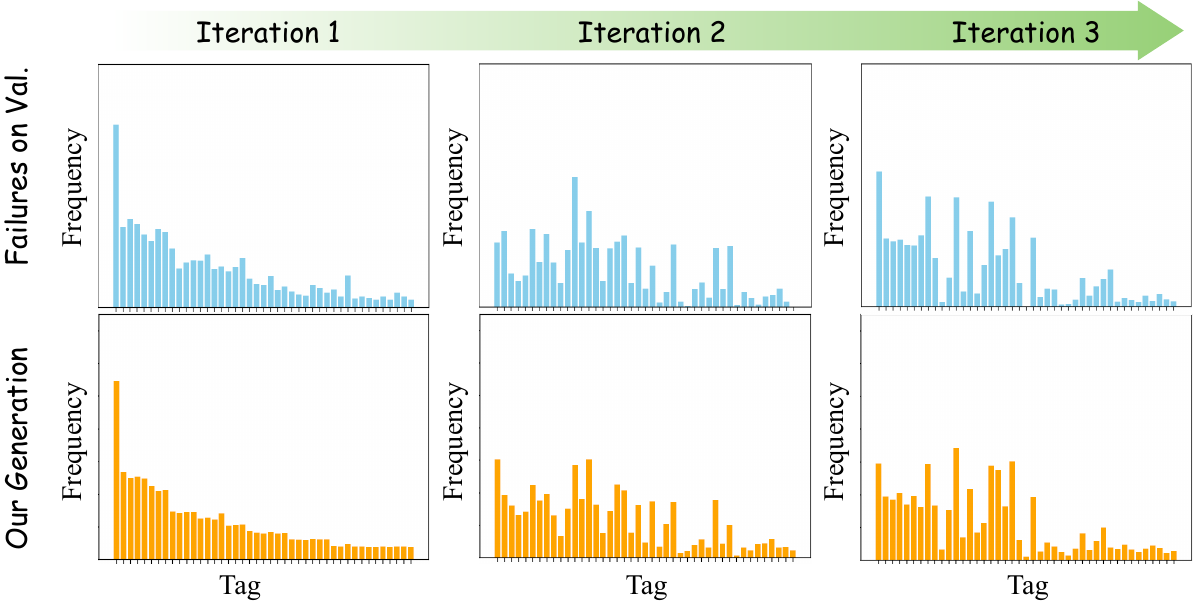}
    \caption{
        \textbf{Distribution gap} between augmented data and failures on the validation set over multi-iterations.
    }
    \vspace{-0.15cm}
    \label{fig:dist_multi_iterations}     
\end{figure}

\mypara{The effect of multiple iterations.}
Our CorrectAD framework is designed as an iterative self-correcting system for E2E models. Within the time constraints, we conducted several cycles of iteration. As indicated in Tab.~\ref{tab:Ablation_multi_iteration}, both the L2 error and collision rate decreased progressively with more iterations. Fig.~\ref{fig:dist_multi_iterations} illustrates the distribution differences between the generated data and the failures in the validation set for each iteration. The visualization demonstrates that, with more iterations, the distribution of the data generated by our method increasingly aligns with the distribution of failures, which explains why our method gradually reduces both the L2 error and collision rate. This highlights the self-correcting potential of our CorrectAD framework.

\section{Conclusion}
In this paper, we propose a self-correcting agentic system, CorrectAD, to effectively improve the E2E models in autonomous driving. We first propose a PM-Agent to analyze failure causes and formulate data requirements. Then, we introduce DriveSora to generate high-fidelity training data, thereby correcting the failures of E2E models. Experiments on multiple datasets proves that CorrectAD shows significant improvements in L2 error and collision rate, showcasing its excellent robustness, and providing a sustainable model self-correction solution for autonomous driving.

\mypara{Limitation and Societal Impact.}
CorrectAD currently only treats collisions as failure cases, omitting issues like lane violations and traffic rule breaches. We plan to broaden this scope using more comprehensive benchmarks~\cite{jia2024bench2drive, dauner2024navsim} that support such evaluations. 
Additionally, CorrectAD employs a powerful diffusion transformer for data generation, but it remains too inefficient for real-time use—DriveSora (1.1B params) requires 8×A800 GPUs for 72h training and 4s per example at inference (L40S). Future work may integrate faster alternatives\cite{xie2024sana}. Overall, CorrectAD shows strong potential for scalable and robust E2E AD systems.

\section{Acknowledgments}
This work is partially supported by the National Natural Science Foundation of China (NSFC) under grant No. 62403389, the Provincial Natural Science Foundation of Zhejiang under grant No. QKWL25F0301, and the Zhejiang Key Laboratory of Low-Carbon Intelligent Synthetic Biology (2024ZY01025).

\bibliography{11_references}


\appendix
\section{Method}
\subsection{PM-Agent}
\begin{figure}[h!]
    \centering
    \includegraphics[width=0.95\linewidth]{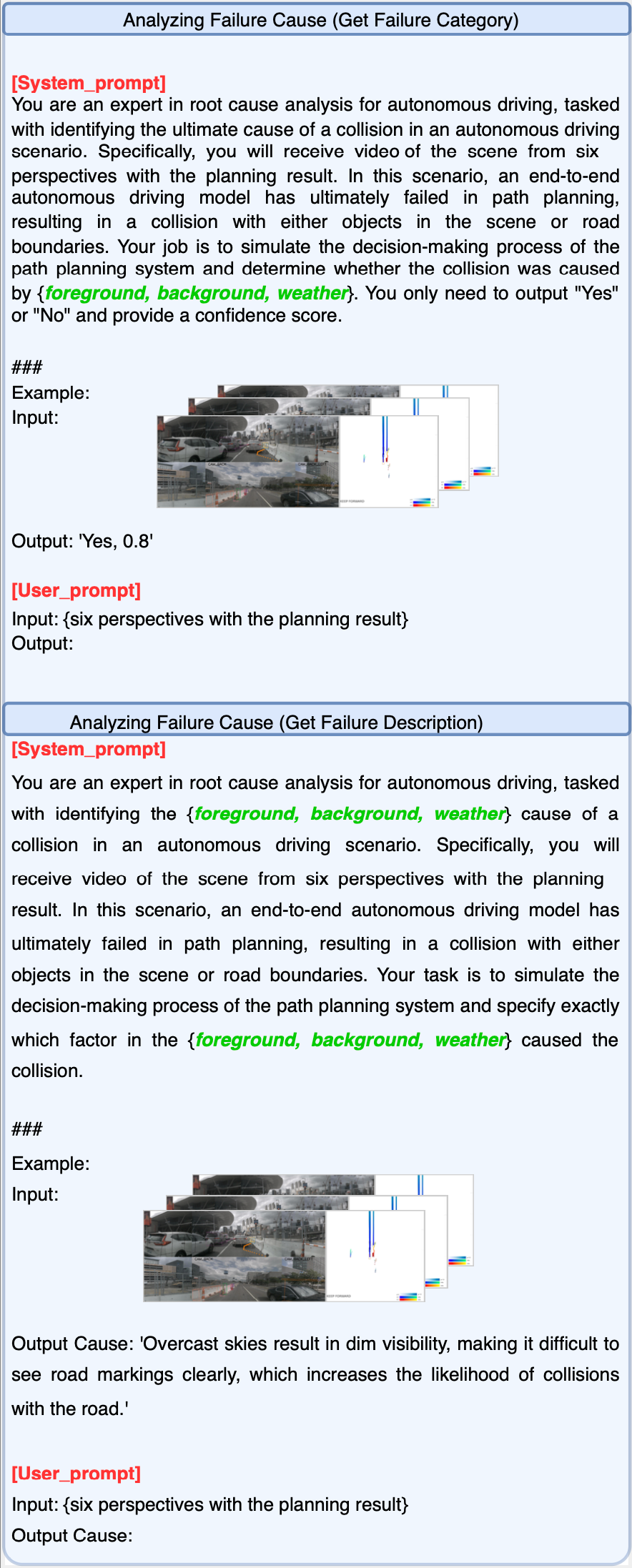}
    \caption{The prompt of \textbf{Analyzing Failure Cause.}}
    \label{fig:prompt_1} 
\end{figure}

\mypara{Analyzing Failure Cause.} 
First, we input the scene images from six perspectives and the path planning results into a VLM. The model assesses potential failure reasons (foreground, background, or weather) three times and outputs the confidence for each category. The prompt used for this process is shown in part (a) of Fig.~\ref{fig:prompt_1}. A threshold of 0.8 is set; if the confidence surpasses this threshold, the VLM is further instructed to provide the specific cause under the identified failure category (foreground, background, or weather). The prompt for this process is shown in part (b) of Fig.~\ref{fig:prompt_1}.

\begin{figure}[h!]
    \centering
    \includegraphics[width=0.95\linewidth]{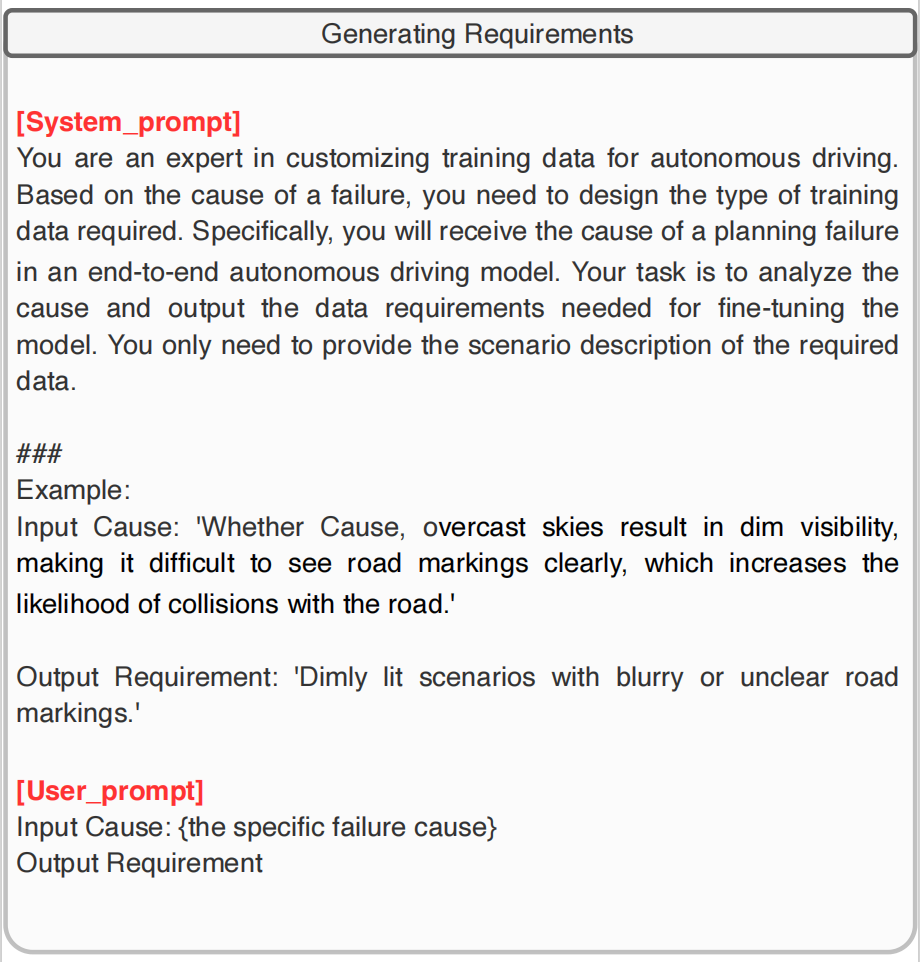}
    \caption{The prompt of \textbf{Generating Requirements.}}
    \label{fig:prompt_2} 
\end{figure}

\mypara{Generating Requirements.} After identifying the specific failure cause, we use an LLM to help summarize precise data requirements based on the identified causes. The prompt used for this step is shown as Fig.~\ref{fig:prompt_2}.

\begin{figure}[h!]
    \centering
    \includegraphics[width=0.95\linewidth]{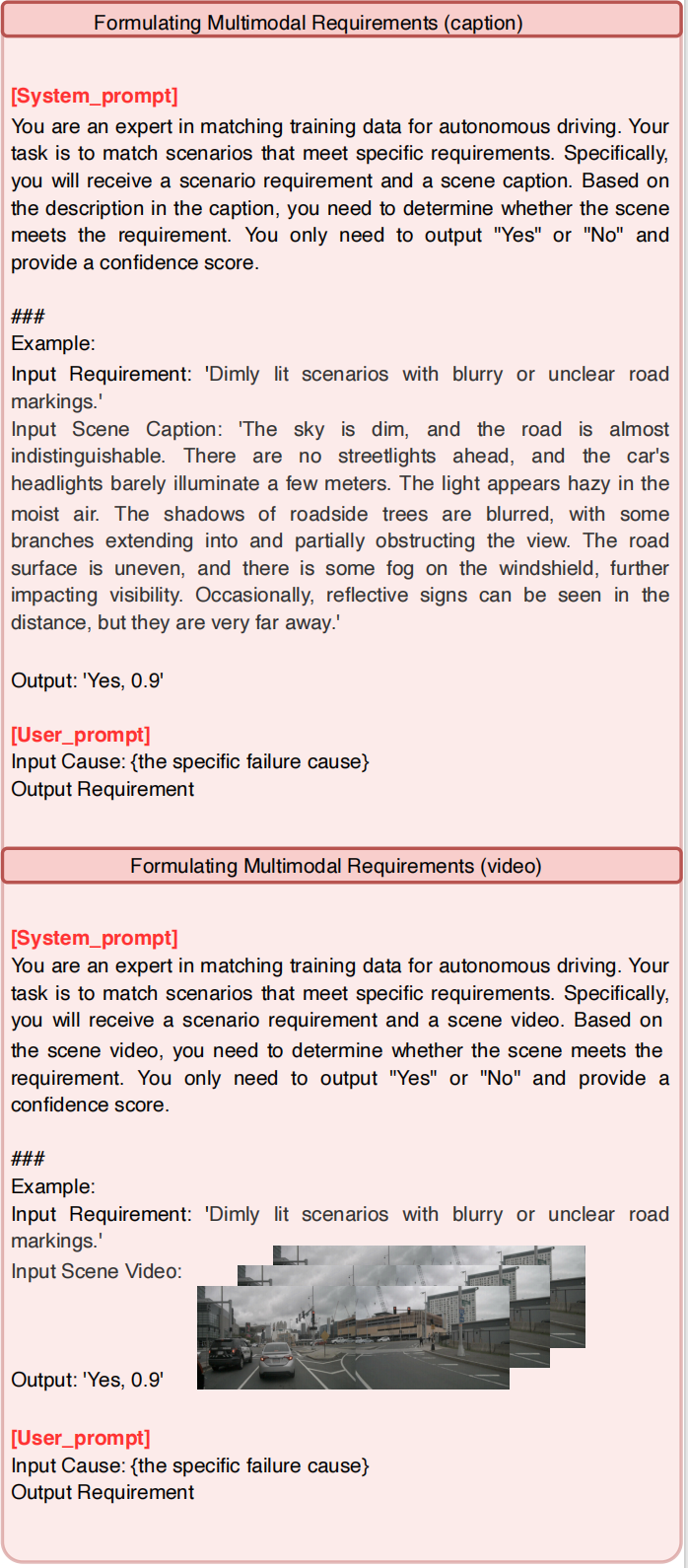}
    \caption{The prompt of \textbf{Formulating Multimodal Requirements.}}
    \label{fig:prompt_3} 
\end{figure}

\mypara{Formulating Multimodal Requirements.} 
After obtaining detailed data requirements, we compare these requirements with the scenarios in the Nuscene dataset. For economic considerations, we sample five evenly spaced frames from each scene for comparison. First, the LLM compares the data requirements with all scene captions for initial screening. The prompt for this process is shown in part (a) of Fig.~\ref{fig:prompt_3}. Next, the VLM compares the data requirements with the images of the remaining scenes from the initial screening, further filtering to identify matching scenes. The prompt for this process is shown in part (b) of Fig.~\ref{fig:prompt_3}. Finally, we extract the captions of the matched scenes along with their corresponding BEV layouts to create a multimodal prompt. This serves as input for the downstream generation model.

\subsection{DriveSora}
\label{supp: drivesora}

\begin{figure}[t]
    \centering
    \includegraphics[width=0.9\linewidth]{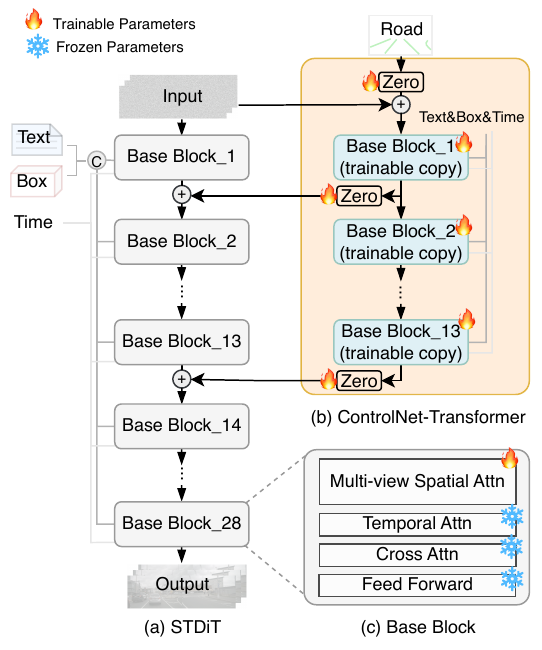}
    \caption{The overview of ControlNet-Transformer in DriveSora.}
    \label{fig:supp_controlnet_transformer} 
\end{figure}

\mypara{ControlNet-Transformer.}
 We illustrate the detailed framework of ControlNet-Transformer in Fig.~\ref{fig:supp_controlnet_transformer}. To introduce road layout conditions into our STDiT network, we follow the ControlNet~\cite{zhang2023_controlnet} by creating a trainable copy of the encoder portion of STDiT. Since Transformer-based models do not have a distinct encoder-decoder structure, following~\cite{chen2024pixart}, we treat the first 13 blocks ($N=13$) of the model as the encoder.
In ControlNet-Transformer, the output of each block passes through a learnable Zero linear layer and is then added to the corresponding block in STDiT. This summed output subsequently serves as the input for the next block.
The integration of ControlNet principles with the Transformer architecture allows for effective conditioning of the model on road layout information. This approach maintains the core functionality of STDiT while enhancing its ability to generate outputs that are consistent with the provided road layout conditions.

\begin{figure}[h!]
    \centering
    \includegraphics[width=0.95\linewidth]{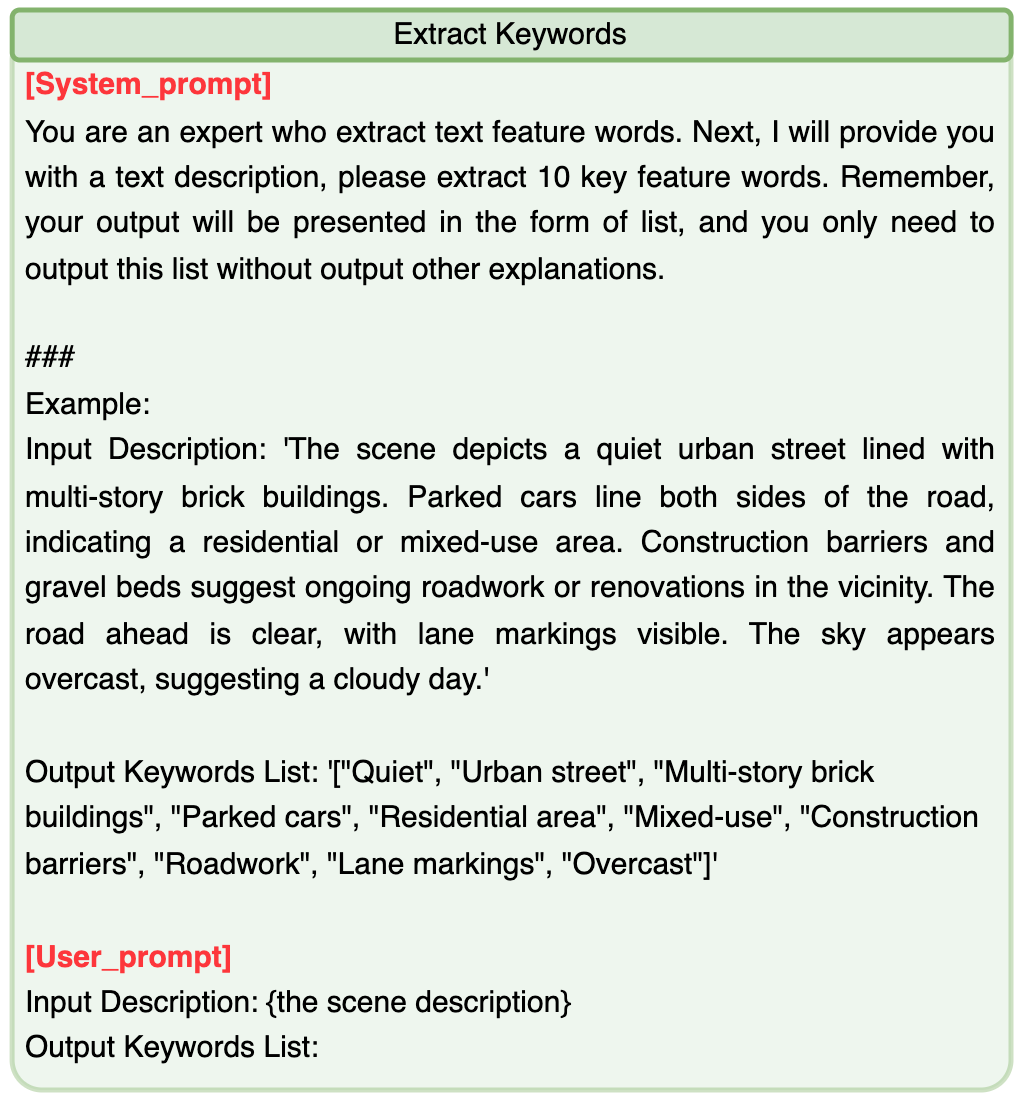}
    \caption{The prompt of \textbf{Extracting Keywords.}}
    \label{fig:prompt_4} 
\end{figure}

\section{Experiments}

    \begin{figure}[t!]
        \centering
    \includegraphics[width=1.0\linewidth]{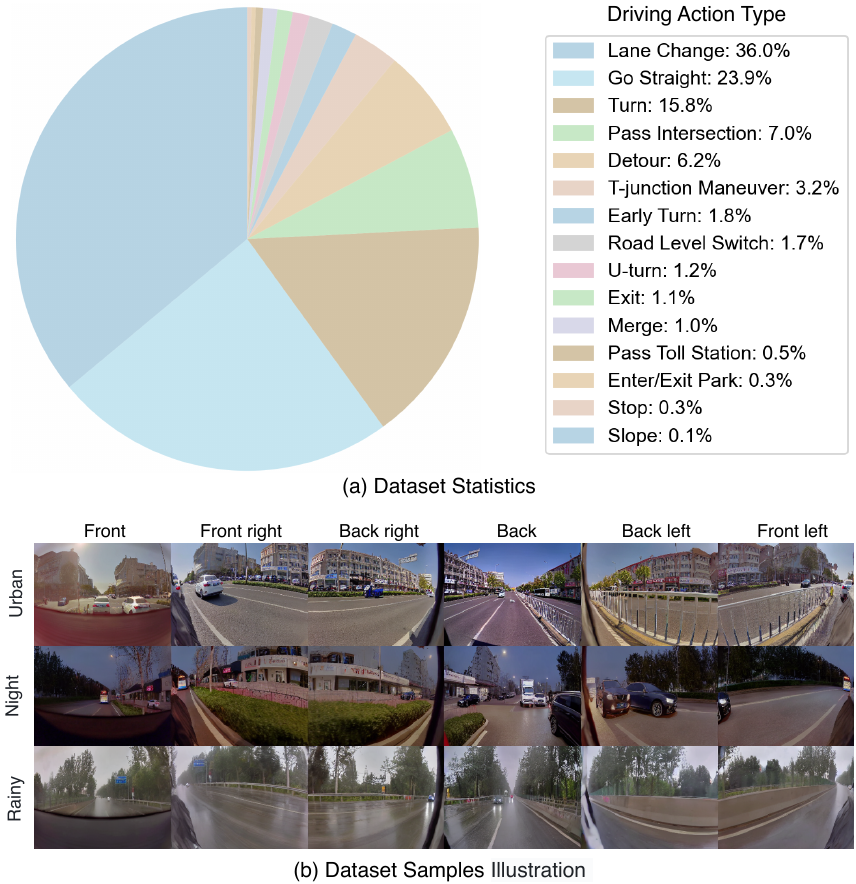}
    \caption{
    Driving action statistics and samples of the in-house E2E dataset.
    }
    \label{fig:supp_liauto_dataset} 
    \end{figure}

\subsection{Dataset}
Considering that the majority of the nuScenes~\cite{caesar2020_nuscenes} dataset consists of relatively simple scenarios (as noted by Ego-MLP~\cite{li2024egomlp}: 73.9\% of the nuScenes data involve scenarios of driving straightforwardly), we further evaluate the effectiveness of CorrectAD on a more challenging in-house dataset. This dataset contains 3.6M samples, which is 3,600 times larger than nuScenes. As illustrated in Fig.~\ref{fig:supp_liauto_dataset}, our in-house dataset exhibits a much richer distribution of driving actions than nuScenes, with lane change being the most common behavior (accounting for 36\%). The scale and complexity of this challenging dataset make our experimental results more convincing and reliable.

\subsection{Metrics}
\label{supp: Metrics}

\mypara{Metrics of the in-house E2E model.}
Our in-house E2E model employs two key metrics: L2 error and Hit Rate. The L2 error metric measures the distance error between the planned trajectory and the recorded trajectory over a time period ranging from 0 seconds to a specified moment. The Hit Rate metric represents the recall rate at a specific time point. It determines whether the planned trajectory points fall within a 3.5-meter diameter around the ground truth trajectory points. The 3.5-meter threshold is chosen because it closely approximates the width of a standard traffic lane. Using both metrics, the model can be evaluated for its continuous trajectory accuracy and point-specific precision, offering a robust assessment of its predictive capabilities in various traffic scenarios.

\subsection{Implementation Details}
\label{supp: Implementation_Details}

\begin{figure}[t]
    \centering
    \includegraphics[width=0.95\linewidth]{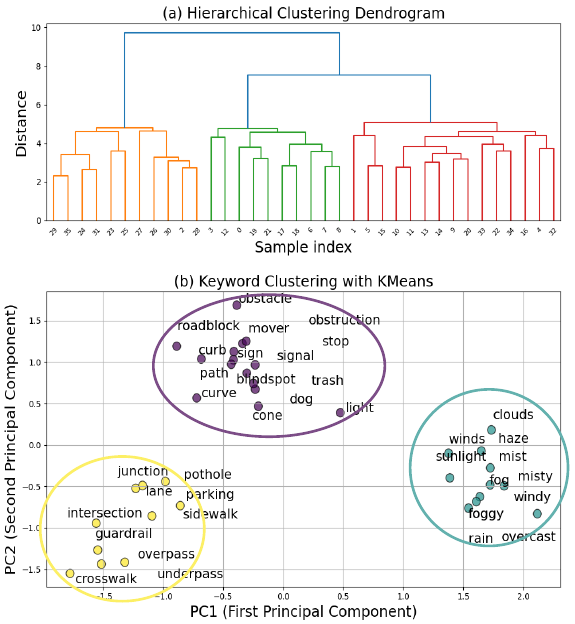}
    \caption{
    \textbf{The clustering result of planning failure.}
    }
    \label{fig:cluster_cause}
\end{figure}

\mypara{Hyperparameters.}
We finetuned the E2E models~\cite{hu2023uniad, Jiang2023VADVS} using combined old and new generated training data, with pre-trained weights and a learning rate of 2e-5. We built PM-Agent based on GPT-4o\footnote{\url{https://platform.openai.com/docs/guides/vision}}. DriveSora, built on OpenSora 1.1~\cite{opensora}, was trained on 8 A800 GPUs.

\mypara{Re-implemented details about AIDE.}
We made several adaptations to AIDE to make it suitable for E2E planning tasks, enabling a fair comparison in this study. Specifically:
(1) Following AIDE’s ``Issue Finder" procedure, we compared the output of the E2E model’s perception module with ground-truth category labels to identify failure categories from failure videos automatically;
(2) Following the ``Data Feeder" stage, we utilized BLIP-2\cite{LI2023blip2} to retrieve samples containing these failure categories from the existing dataset, based on image-text similarity;
(3) Following the ``Pseudo-Labeling" step, since 2D detectors are not applicable to 3D tasks, we adopted a popular 3D auto-labeling detector\cite{wang2023streampetr} to produce 3D bounding box labels, while using expert trajectories as the ground truth for planning;
(4) Following the ``Continual Training" process, we combined the newly assembled samples with the original dataset to further fine-tune the E2E model.

\mypara{Failure cause annotation.}
Firstly, we extracted 27 failure cases from the first training (\textit{i.e.}, $N_\text{anno}=27$). To ensure the accuracy and professionalism of the annotations, we hired domain experts to manually annotate these 27 scenes, providing 10 failure reason annotations for each case. Subsequently, an anonymous cross-voting method was used to select the top 3 annotation results from experts, ensuring the objectivity and effectiveness of the annotation process.

\mypara{Clustering of the failure categories.}
We employed GPT-4o to extract keywords from the annotated failure reasons and performed fuzzy clustering on all extracted keywords to merge similar terms, such as ``rain" and ``rainy." During this process, an Euclidean distance threshold of 0.8 was set, resulting in 32 keywords. Then, hierarchical clustering was applied to analyze these 32 keywords, and the resulting dendrogram is shown in Fig.~\ref{fig:cluster_cause}(a). Based on the clustering results, dividing the keywords into 3 clusters was determined to be the optimal choice, so $K=3$ was selected. Next, we used the K-means clustering algorithm to categorize all keywords into three groups, with the clustering results presented in Fig.~\ref{fig:cluster_cause}(b). Finally, we input these three groups of keywords into GPT-4o and asked it to generate a label for each category, resulting in the three labels ``Foreground,'' ``Background,'' and ``Weather.''

\mypara{Training and inference details of DriveSora.}
The original image size in nuScenes is 1600x900. We resize these images to 512x512 for model training. Initially, we fine-tune a single-view video model on nuScenes. This model uses multimodal prompts as conditions, including scene descriptions and BEV layouts. We first project the BEV layout onto the camera perspective, resulting in 3D bounding boxes and road sketches. For discrete box conditions, we concatenate them with scene descriptions along the token dimension and inject them into the cross-attention layer. For road sketches, we incorporate them into the original STDiT network using a trainable ControlNet-Transformer.
We initialize the single-view video model using the checkpoint from OpenSora 1.1~\cite{opensora}, with a video frame length T=16. This single-view video model is trained for 30,000 iterations with a total batch size of 16. We employ the HybridAdam optimizer with a learning rate of 2e-5. Subsequently, we modify the spatial self-attention parameters to construct a multi-view video model. This multi-view video model is trained for 25,000 iterations with a total batch size of 16, using the HybridAdam optimizer with a learning rate of 2e-5. For CFG during training, each condition has 5\% probability to be set as null $\phi$, with another 5\% chance of setting all to $\phi$.

For inference, we employ rectified flow sampling with 30 steps. We utilize classifier-free guidance (CFG) to enhance conditional guidance. The values for $\lambda_T, \lambda_B$, and $\lambda_R$ are set to 2.0, 2.0, and 7.0, respectively. Each inference generates a 16-frame video sequence. Similar to methods~\cite{blattmann2023align_videoldm, wang2023driveWM}, we utilize the last 4 frames of the generated video as conditions for subsequent long video generation.

\mypara{Modeling approach of the statistical distribution.} First, we use LLM to extract keywords from the captions of all scenes in the Nuscene dataset. The specific prompt is shown in the Fig.~\ref{fig:prompt_4}. Next, we perform fuzzy clustering on all extracted keywords, with the Euclidean distance threshold set to 0.8. Finally, we select the top 100 most frequently occurring keywords as labels, arranged in order of frequency. We then compute the occurrence frequency of these labels across different datasets and plot the distribution, with the horizontal axis representing the labels and the vertical axis representing the frequency.

\subsection{Additional Visual Results}
\label{supp: Additional_Visual_Results}

\begin{figure*}[t!]
    \centering
    \includegraphics[width=0.9\linewidth]{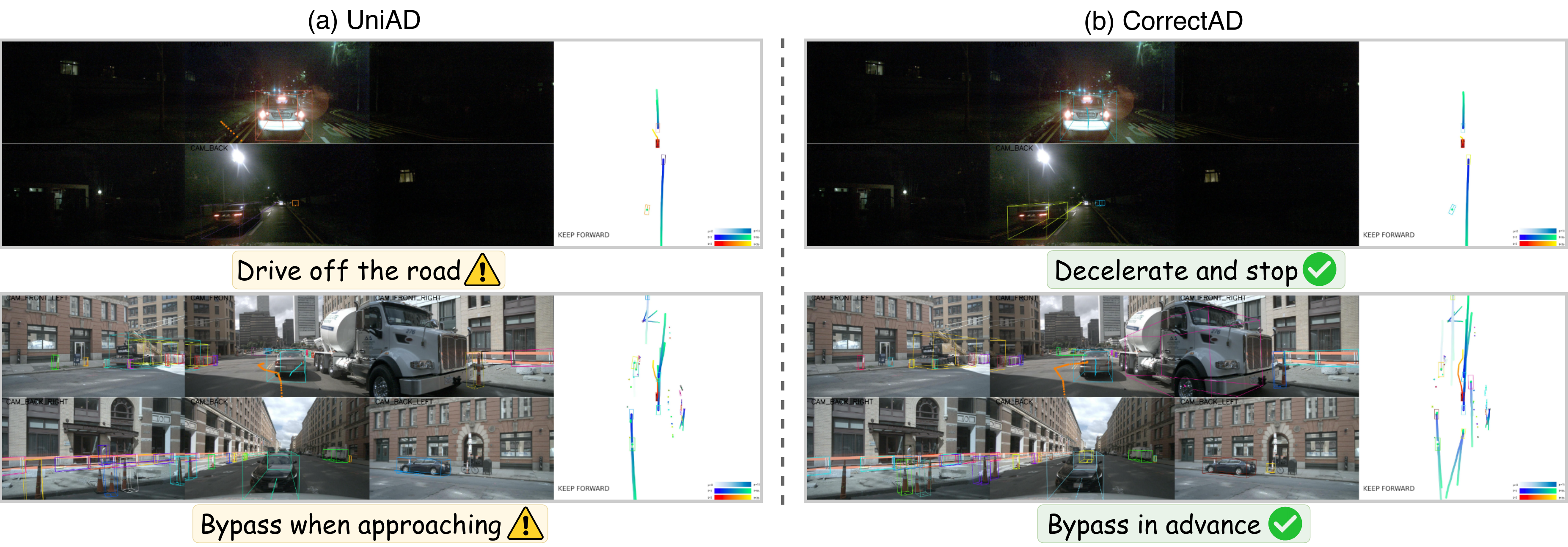}
    \caption{
    \textbf{Visualization of two examples before and after self-correction on nuScenes validation set.} \textbf{(a)} We show two hard examples from the validation set, ``a low-visibility night'', ``bypass in dense traffic flow''. \textbf{(b)} Our framework can fix these examples. 
    }
    \label{fig:supp:uniad generalization comparison} 
\end{figure*}

\mypara{Failure case corrections on nuScenes dataset.} 
In Fig.~\ref{fig:supp:uniad generalization comparison}, we present two examples before and after self-correction on the nuScenes validation set, with all six camera views and one BEV view output by UniAD.

\begin{figure*}[t]
    \centering
    \includegraphics[width=0.95\linewidth]{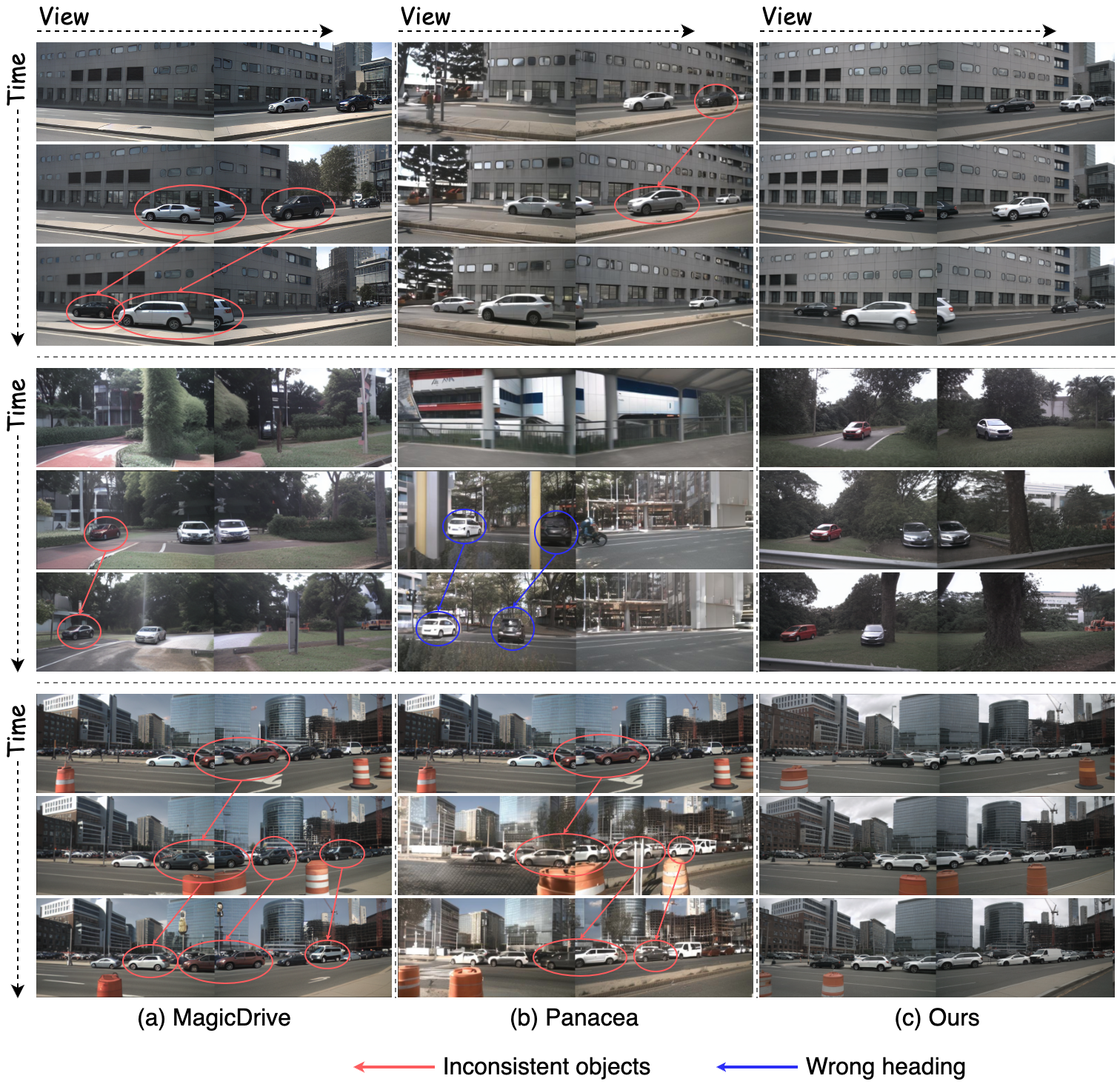}
    \caption{\textbf{Visual comparison of different generative models}. Our DriveSora maintains consistent spatial-temporal appearance where the previous methods fail.}
    \label{fig:supp visual conparison of local region} 
\end{figure*}

\mypara{Comparison of results from different generative models.} Fig.~\ref{fig:supp visual conparison of local region} shows more visual comparison of local region generated by different generative models. This indicates that the foreground objects generated by our method maintain superior consistency over time.

\begin{figure*}[t]
    \centering
      \includegraphics[width=0.97\linewidth]{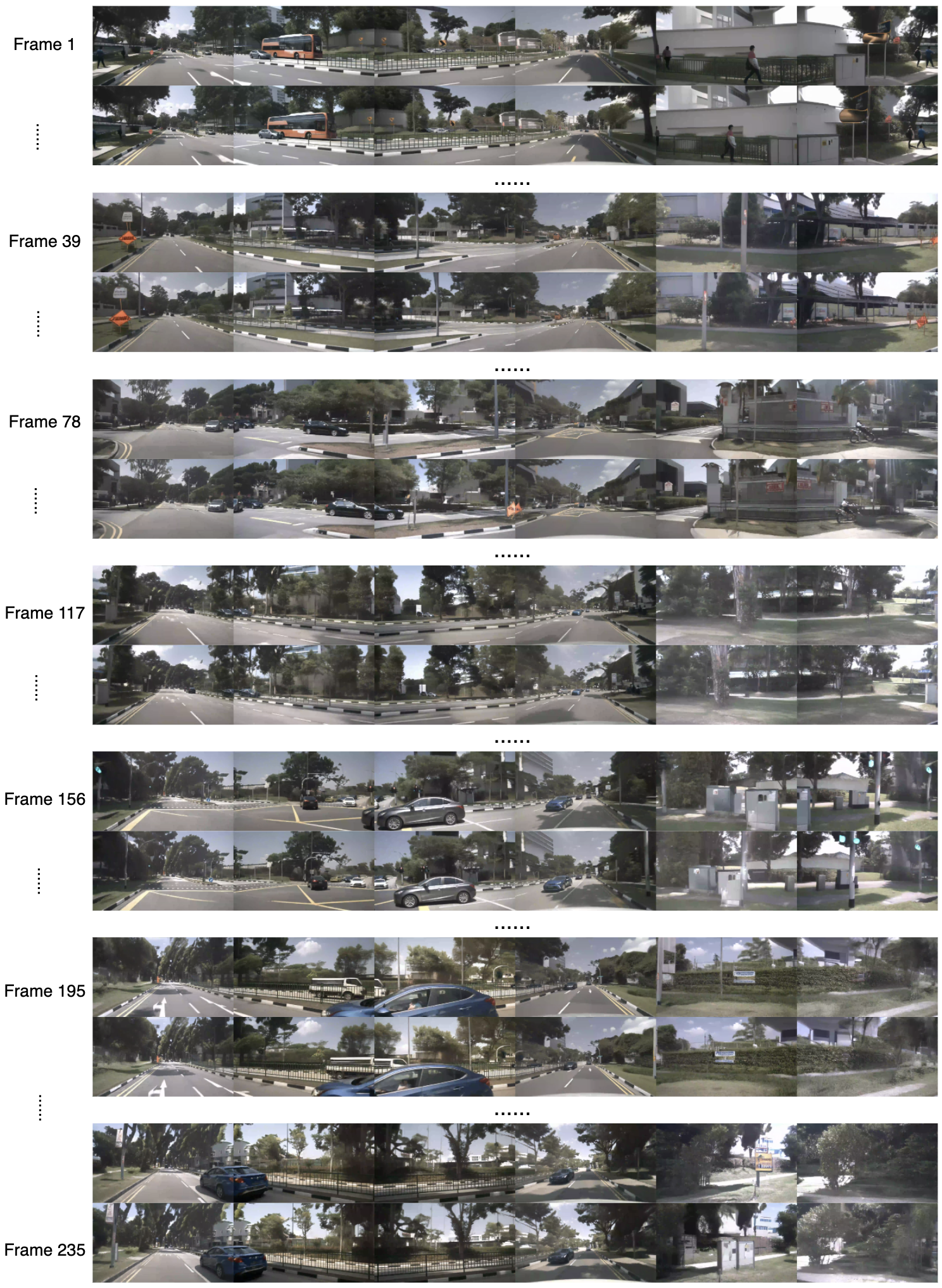}
    \caption{
    {The multi-view video generated by DriveSora on the nuScenes validation set.}
    }
    \label{fig:supp_generation_on_nusc} 
\end{figure*}

\begin{figure*}[t]
    \centering
      \includegraphics[width=0.97\linewidth]{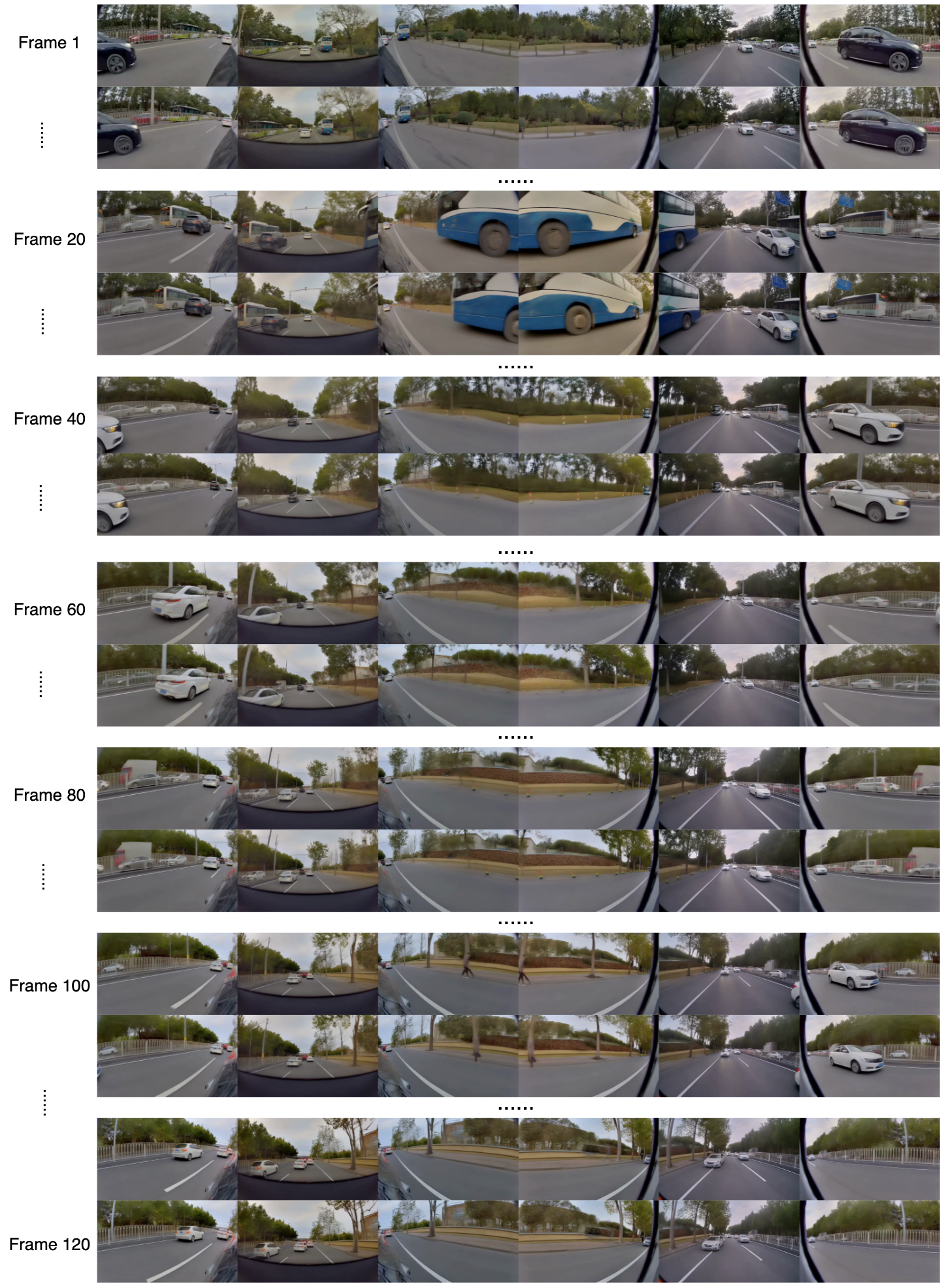}
    \caption{
    {The multi-view video generated by DriveSora on the in-house dataset.}
    }
    \label{fig:supp_generation_on_inhouse} 
\end{figure*}

\begin{figure*}[t!]
    \centering
      \includegraphics[width=0.95\linewidth]{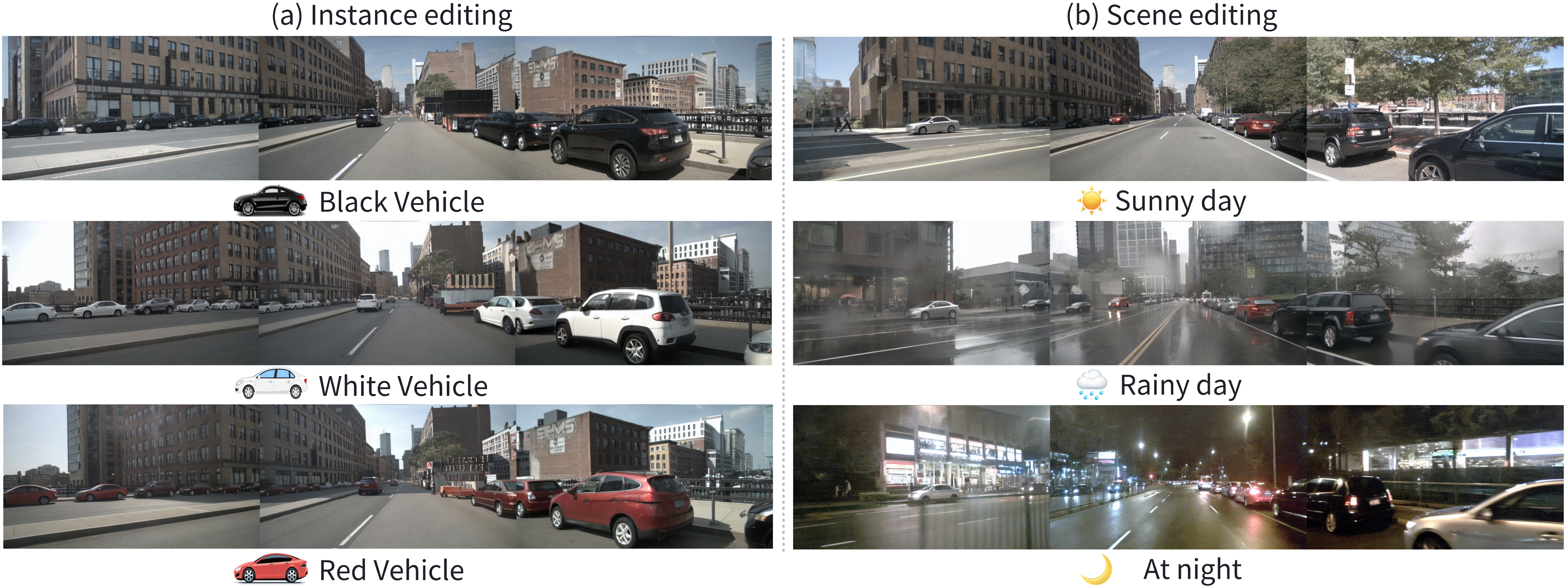}
    \caption{
    \textbf{Visualization of instance and scene editing. }
    \textbf{(a)} shows the instance-level control result, such as the appearance attributes of all vehicles. \textbf{(b)} shows the scene-level control result, including weather and time.}
    \label{fig:visual precise control} 
\end{figure*}

\mypara{Multi-view video generation on multiple datasets.} In Fig.~\ref{fig:supp_generation_on_nusc} and~\ref{fig:supp_generation_on_inhouse}, we present the multi-view video generated by our DriveSora using the nuScenes dataset and our in-house dataset, respectively. The generated video maintains perfect spatial and temporal consistency. In addition, Fig.~\ref{fig:visual precise control} shows that DriveSora can flexibly control the properties of the foreground vehicle and the background weather.

\subsection{Additional Analysis}
\mypara{Ablation of DriveSora.}
Tab.~\ref{tab:supp_ablation_of_drivesora} illustrates the ablation study of the Multimodal Prompt and Multi-view Spatial Attention in DriveSora. When neither component is used, the score reaches the worst. Incorporating only the Multimodal prompt significantly improves the scores, especially in NDS, which rises to 36.37. The optimal setup utilizes both components, leading to the lowest FID of 15.08 and the highest CLIP and NDS scores of 86.73 and 36.58, respectively, demonstrating the complementary effects of these features in enhancing DriveSora's performance.

\setlength{\tabcolsep}{3.5pt}
\begin{table}[]
\centering
\small
\begin{tabular}{c|c|cccc}
\toprule
\textbf{\begin{tabular}[c]{@{}c@{}}Multimodal\\ Prompt\end{tabular}} & \textbf{\begin{tabular}[c]{@{}c@{}}Multi-view\\ Spatial Attn\end{tabular}} & \textbf{FID $\downarrow$} & \textbf{CLIP $\uparrow$} & \textbf{FVD $\downarrow$} & \textbf{NDS $\uparrow$} \\
\midrule
{\color[HTML]{C0C0C0} \XSolidBrush} & {\color[HTML]{C0C0C0} \XSolidBrush} & 25.65 & 72.28 & 97.32 & 25.23 \\
\Checkmark & {\color[HTML]{C0C0C0} \XSolidBrush} & 17.23 & 79.5 & 95.18 & 36.37 \\
\rowcolor{gray!15} 
\Checkmark & \Checkmark & \textbf{15.08} & \textbf{86.73} & \textbf{94.51} & \textbf{36.58} \\
\bottomrule
\end{tabular}
\caption{Ablation of the Multimodal prompt and Multi-view Spatial Attn in DriveSora. }
\label{tab:supp_ablation_of_drivesora}
\end{table}

\noindent \textbf{Ablation of Classifier-free Guidance.}
We compared various CFG methods, considering both conditional and unconditional foreground and background elements, as summarized in Tab.~\ref{tab:supp_CFG}. Our proposed method, CFG$_{Text,Fore,Back}$, was evaluated alongside other approaches. When we excluded the unconditional sketch (CFG$_{Text,Fore}$) or both sketch and background (CFG$_{Text}$), we observed slightly better FVD scores, but these configurations exhibited more significant differences in BEV segmentation and 3D object detection. Additionally, we tested CFG$_{MagicDrive}$ from MagicDrive \cite{gao2023magicdrive}, which performed well in terms of controllability but showed only satisfactory FVD. In conclusion, CFG$_{Text,Fore,Back}$ achieved the best overall performance across all evaluated criteria.

\setlength{\tabcolsep}{3.5pt}
\begin{table}[]
\centering
\begin{tabular}{c|ccc}
\toprule
Method          & FVD$_\downarrow$   & Object mAP$_\uparrow$ & Map mIoU$_\uparrow$   \\ \midrule
CFG$_{Text,Fore,Back}$      & 94.60 & 24.55      & \textbf{35.96}    \\
CFG$_{Text,Fore}$        & 89.12 & 24.70      & 34.40     \\
CFG$_{Text}$          & \textbf{83.63} &  20.05          & 34.26        \\
CFG$_{MagicDrive}$ & 164.48 & \textbf{26.18}     & 35.02    \\ \bottomrule
\end{tabular}
\caption{Ablation on the classifier-free guidance.}
\label{tab:supp_CFG}
\end{table}

\mypara{Closed‑loop Evaluation.}
As shown below, CorrectAD achieves a 0.9 PDMS improvement over LTF baseline~\cite{chitta2022transfuser} on the NAVSIM navtest~\cite{dauner2024navsim} closed-loop benchmark, indicating better planning robustness.

\setlength{\tabcolsep}{2pt}
\begin{table}[h]
\centering
\begin{tabular}{l|cccccc}
\toprule
Method & NC$\uparrow$ & DAC$\uparrow$ & TTC$\uparrow$ & Comf.$\uparrow$ & EP$\uparrow$ & PDMS$\uparrow$ \\
\midrule
LTF        & 97.4 & 92.8 & 92.4 & 100 & 79.0 & 83.8 \\
+CorrectAD    & 98.0 & 93.2 & 93.3 & 100 & 79.3 & 84.7 (+0.9) \\
\bottomrule
\end{tabular}
\caption{Closed‑loop results on NAVSIM navtest benchmark.}
\end{table}

\mypara{Case study of CorrectAD.}
Tab.~\ref{tab:supp_casestudy} presents a case study of failure scenarios concerning UniAD in the nuScenes validation set over three iterations. It tracks the model's ability to address previously unresolved cases and handle new failures.
Initially, there were 22 total failures (Iteration 0). Throughout the iterations, the number of old unresolved cases decreases, and by Iteration 3, the model reduces the total failures to 14 with 10 unresolved old cases and 4 new ones.
This demonstrates a model improvement with a 62.5\% error resolution rate. The rate of new errors (``forgetting rate") remains within a manageable range, indicating effective model updates. With more iterations, it's hopeful that the model will get even stronger and more adaptable, leading to better accuracy and reliability in future model versions.

\setlength{\tabcolsep}{10pt}
\begin{table}[]
\centering
\begin{tabular}{c|ccc }
\toprule
\multicolumn{1}{l|}{\textbf{Iteration}} & \textbf{Old $\downarrow$} & \textbf{New $\downarrow$} & \textbf{Total $\downarrow$} \\
\midrule
0 & - & - & 22 \\
1 & 18 & 1 & 19 \\
2 & 13 & 3 & 16 \\
3 & 10 & 4 & 14 \\ 
\bottomrule
\end{tabular}
\caption{Case study of failure scenarios about UniAD~\cite{hu2023uniad} in the nuScenes validation set. ``Total'' refers to the total number of failures. ``Old'' indicates the number of unresolved cases from the previous iteration's failure set. ``New'' refers to newly failed cases not part of the previous iteration's failure set.}
\label{tab:supp_casestudy}
\end{table}

\end{document}